%% file: main.tex
\documentclass[sigconf]{acmart}

\pdfoutput=1
\makeatletter
\def\mdseries@tt{m}
\makeatother 

\usepackage{graphicx}
\usepackage{amsmath, amsfonts, amssymb, amsthm} 
\usepackage{hyperref}           
\usepackage{tikz}
\usepackage[
font=Large
]{subfig}
\usetikzlibrary{arrows,petri,topaths,backgrounds,patterns,positioning}
\usepackage{pgfplots}
\usepgfplotslibrary{groupplots}
\usepackage{units}
\usepackage{booktabs}
\usepackage{tabularx}
\usepackage{algorithmic}
\usepackage[linesnumbered,lined,boxed]{algorithm2e}

\usepgfplotslibrary{fillbetween}

\AtBeginDocument{%
	\providecommand\BibTeX{{%
			\normalfont B\kern-0.5em{\scshape i\kern-0.25em b}\kern-0.8em\TeX}}}

\newcommand{\myparagraph}[1]{\vspace*{1mm} {\noindent\bf #1} }

\newcommand{\specialcell}[2][c]{%
	\begin{tabular}[#1]{@{}c@{}}#2\end{tabular}}

\copyrightyear{2020}
\acmYear{2020}
\setcopyright{acmcopyright}\acmConference[CIKM '20]{Proceedings of the 29th ACM International Conference on Information and Knowledge Management}{October 19--23, 2020}{Virtual Event, Ireland}
\acmBooktitle{Proceedings of the 29th ACM International Conference on Information and Knowledge Management (CIKM '20), October 19--23, 2020, Virtual Event, Ireland}
\acmPrice{15.00}
\acmDOI{10.1145/3340531.3411866}
\acmISBN{978-1-4503-6859-9/20/10}

\newtheorem{remark}{Remark}
\newtheorem{definition}{Definition}

\begin{document}

\title{Characteristic Functions on Graphs: Birds of a Feather, from Statistical Descriptors to Parametric Models}


\author{Benedek Rozemberczki}
\affiliation{%
  \institution{The University of Edinburgh}
  \city{Edinburgh}
  \country{United Kingdom}}
\email{benedek.rozemberczki@ed.ac.uk}

\author{Rik Sarkar}
\affiliation{%
  \institution{The University of Edinburgh}
  \city{Edinburgh}
  \country{United Kingdom}}
\email{rsarkar@inf.ed.ac.uk}

\begin{abstract}
In this paper, we propose a flexible notion of characteristic functions defined on graph vertices to describe the distribution of vertex features at multiple scales. We introduce \textit{FEATHER}, a computationally efficient algorithm to calculate a specific variant of these characteristic functions where the probability weights of the characteristic function are defined as the transition probabilities of random walks. We argue that features extracted by this procedure are useful for node level machine learning tasks. We discuss the pooling of these node representations, resulting in compact descriptors of graphs that can serve as features for graph classification algorithms. We analytically prove that \textit{FEATHER} describes isomorphic graphs with the same representation and exhibits robustness to data corruption. Using the node feature characteristic functions we define parametric models where evaluation points of the functions are learned parameters of supervised classifiers. Experiments on real world large datasets show that our proposed algorithm creates high quality representations, performs transfer learning efficiently, exhibits robustness to hyperparameter changes and scales linearly with the input size.
\end{abstract}

\maketitle

\input{./sections/introduction.tex}

\input{./sections/related_work.tex}

\input{./sections/characteristic_function.tex}

\input{./sections/experiments.tex}

\input{./sections/conclusions.tex}

\section*{Acknowledgements}
Benedek Rozemberczki was supported by the Centre for Doctoral Training in Data Science, funded by EPSRC (grant EP/L016427/1). We would also like to thank Rayna Andreeva, Maria Astefanoaei, Tyler Derr, Abhirup Ghosh, Anupam Mediratta and Lauren Watson for their useful comments.

\bibliographystyle{ACM-Reference-Format}

\bibliography{main}

\end{document}

%% file: sections/introduction.tex



\section{Introduction}

Recent works in network mining have focused on characterizing node neighbourhoods. Features of a neighbourhood serve as valuable inputs to downstream machine learning tasks such as node classification, link prediction and community detection~\cite{deepwalk, grarep, node2vec, rolx, tadw}. In social networks, the importance of neighbourhood features arises from the property of homophily (correlation of network connections with similarity of attributes), and social neighbours have been shown to influence habits and attributes of individuals~\cite{perozzi2015exact}. Attributes of a neighbourhood is found to be important in other types of networks as well. Network mining methods have used aggregate features from several degrees of neighbourhoods for network analysis and embedding~\cite{bane,sine,tadw,rolx}. 

Neighbourhood features can be complex to interpret. Network datasets can incorporate multiple attributes, with varied distributions that influence the characteristics of a node and the network. Attributes such as income, wealth or number of page accesses can have an unbounded domain, with unknown distributions. Simple linear aggregates~\cite{bane,sine,tadw,rolx} such as the mean values do not represent this diverse information. 


We use characteristic functions~\cite{lukacs} as a rigorous but versatile way of utilising diverse neighborhood information.  A unique characteristic function always exists irrespective of the nature of the distribution, and characteristic functions can be meaningfully composed across multiple nodes and even multiple attributes. These features let us represent and compare different neighborhoods in a unified framework.





\input{./figures/wiki_explanatory.tex}

Figure \ref{fig:selling} shows the distribution of node level characteristic function values on the Wikipedia Crocodiles web graph \cite{musae}. In this dataset nodes are webpages which have two types of labels -- popular and unpopular. With log transformed degree centrality as the vertex attribute, we conditioned the distributions on the class memberships. We plotted the mean of the distribution at each evaluation point with the standard deviation around the mean. One can easily observe that the value of the characteristic function is discriminative with respect to the class membership of nodes. Our experimental results about node classification in Subsection~\ref{subsec:node_classification} validates this observation about characteristic functions for the Wikipedia dataset and various other social networks.


\myparagraph{Present work.} We propose complex valued characteristic functions~\cite{lukacs} for representation of neighbourhood feature distributions. Characteristic functions are analogues of  Fourier Transforms defined for probability distributions. We show that these continuous functions can be evaluated suitably at discrete points to obtain effective characterisation of neighborhoods and  describe an approach to learn the appropriate evaluation points for a given task. 


The correlation of attributes are known to decrease with the decrease in tie strength, and with increasing distance between nodes~\cite{cohen2014distance}. We use a random-walk based tie strength definition, where tie strength at the scale $r$ between a source and target node pair is the probability of an $r$ length random walk from the source node ending at the target. We define the \textit{r-scale random walk weighted characteristic function} as the characteristic function weighted by these tie strengths. We propose \textit{FEATHER} an algorithm to efficiently evaluate this function for multiple features on a graph. 

We theoretically prove that graphs which are isomorphic have the same pooled characteristic function when the mean is used for pooling node characteristic functions. We argue that the \textit{FEATHER} algorithm can be interpreted as the forward pass of a parametric statistical model (e.g. logistic regression or a feed-forward neural network). Exploiting this we define the $r$-scale random walk weighted characteristic function based softmax regression and graph neural network models (respectively named \textit{FEATHER-L} and \textit{FEATHER-N}).
 
We evaluate \textit{FEATHER} model variants on two machine learning tasks -- node and graph classification. Using data from various real world social networks (Facebook, Deezer, Twitch) and web graphs (Wikipedia, GitHub), we compare the performance of \textit{FEATHER} with graph neural networks, neighbourhood preserving and attributed node embedding techniques. Our experiments illustrate that \textit{FEATHER} outperforms comparable unsupervised methods by as much as 4.6\% on node labeling and 12.0\% on graph classification tasks in terms of test AUC score. 
The proposed procedures show competitive transfer learning capabilities on social networks and the supervised \textit{FEATHER} variants show a considerable advantage over the unsupervised model, especially when the number of evaluation points is limited. Runtime experiments establish that \textit{FEATHER} scales linearly with the input size.

\myparagraph{Main contributions.} To summarize, our paper makes the following contributions:
\begin{enumerate}
    \item We introduce a generalization of characteristic functions to node neighbourhoods, where the probability weights of the characteristic function are defined by tie strength.
    \item We discuss a specific instance of these functions -- the r-scale random walk weighted characteristic function. We propose  \textit{FEATHER}, an algorithm that calculates these characteristic functions efficiently to create Euclidean node embeddings.
    \item We demonstrate that this function can be applied simultaneously to multiple features. 
    \item We show that the $r$-scale random walk weighted characteristic function calculated by \textit{FEATHER} can serve as the building block for an end-to-end differentiable parametric classifier.
    \item We experimentally assess the behaviour of \textit{FEATHER} on real world node and graph classification tasks.
\end{enumerate}
The remainder of this work has the following structure. In Section \ref{sec:feather_related_work} we overview
the relevant literature on node embedding techniques, graph kernels and neural networks. We introduce characteristic functions defined on graph vertices in Section \ref{sec:feather_theory} and discuss using them as building blocks in parametric statistical models. We empirically evaluate \textit{FEATHER} on various node and graph classification tasks, transfer learning problems, and test its sensitivity to hyperparameter changes in Section \ref{sec:feather_experiments}. The paper concludes with Section \ref{sec:feather_conclusions} where we discuss our main findings and point out directions for future work. The newly introduced node classification datasets and a Python reference implementation of \textit{FEATHER} is available at \url{https://github.com/benedekrozemberczki/FEATHER}.

%% file: figures/wiki_explanatory.tex
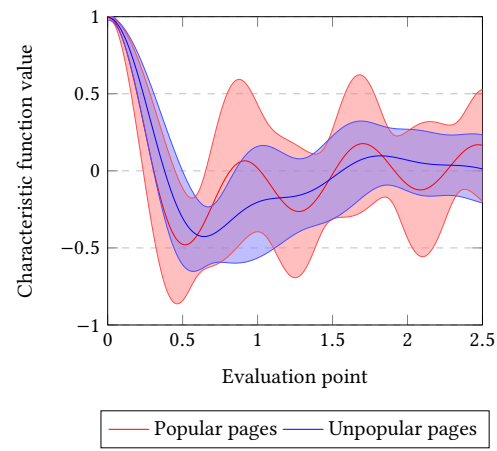
\begin{figure}[h!]
\scalebox{0.9}{
\begin{tikzpicture}
    \begin{axis}[
      xmin=0,
      xmax=2.5,
      ymin=-1,
      ymax=1,
      xtick={0,0.5,1,1.5,2,2.5},
      legend style={at={(.5,-0.4)},anchor=south},
	  legend columns=-1,
      xlabel ={Evaluation point},
      ylabel ={Characteristic function value},
      ymajorgrids=true,
      grid style=dashed,
      width=0.4\textwidth,]

\addplot[color=red] coordinates {(0.005,0.9979)
(0.01,0.9968)
(0.015,0.995)
(0.02,0.9924)
(0.025,0.989)
(0.03,0.9849)
(0.035,0.9801)
(0.04,0.9746)
(0.045,0.9683)
(0.05,0.9613)
(0.055,0.9537)
(0.06,0.9453)
(0.065,0.9362)
(0.07,0.9265)
(0.075,0.9161)
(0.08,0.905)
(0.085,0.8933)
(0.09,0.8809)
(0.095,0.868)
(0.1,0.8544)
(0.105,0.8402)
(0.11,0.8255)
(0.115,0.8102)
(0.12,0.7943)
(0.125,0.778)
(0.13,0.7611)
(0.135,0.7437)
(0.14,0.7259)
(0.145,0.7076)
(0.15,0.6889)
(0.155,0.6697)
(0.16,0.6502)
(0.165,0.6303)
(0.17,0.61)
(0.175,0.5894)
(0.18,0.5685)
(0.185,0.5473)
(0.19,0.5258)
(0.195,0.5041)
(0.2,0.4822)
(0.205,0.4601)
(0.21,0.4378)
(0.215,0.4153)
(0.22,0.3927)
(0.225,0.37)
(0.23,0.3473)
(0.235,0.3244)
(0.24,0.3015)
(0.245,0.2786)
(0.25,0.2557)
(0.255,0.2329)
(0.26,0.2101)
(0.265,0.1873)
(0.27,0.1647)
(0.275,0.1422)
(0.28,0.1198)
(0.285,0.0976)
(0.29,0.0756)
(0.295,0.0537)
(0.3,0.0321)
(0.305,0.0108)
(0.31,-0.0103)
(0.315,-0.0312)
(0.32,-0.0517)
(0.325,-0.0719)
(0.33,-0.0917)
(0.335,-0.1112)
(0.34,-0.1303)
(0.345,-0.1491)
(0.35,-0.1674)
(0.355,-0.1853)
(0.36,-0.2028)
(0.365,-0.2198)
(0.37,-0.2364)
(0.375,-0.2525)
(0.38,-0.2681)
(0.385,-0.2832)
(0.39,-0.2978)
(0.395,-0.3118)
(0.4,-0.3254)
(0.405,-0.3384)
(0.41,-0.3509)
(0.415,-0.3628)
(0.42,-0.3741)
(0.425,-0.3849)
(0.43,-0.3951)
(0.435,-0.4048)
(0.44,-0.4138)
(0.445,-0.4223)
(0.45,-0.4302)
(0.455,-0.4375)
(0.46,-0.4443)
(0.465,-0.4504)
(0.47,-0.4559)
(0.475,-0.4609)
(0.48,-0.4653)
(0.485,-0.4691)
(0.49,-0.4723)
(0.495,-0.475)
(0.5,-0.4771)
(0.505,-0.4786)
(0.51,-0.4796)
(0.515,-0.48)
(0.52,-0.4799)
(0.525,-0.4793)
(0.53,-0.4781)
(0.535,-0.4764)
(0.54,-0.4743)
(0.545,-0.4716)
(0.55,-0.4684)
(0.555,-0.4648)
(0.56,-0.4607)
(0.565,-0.4561)
(0.57,-0.4511)
(0.575,-0.4457)
(0.58,-0.4399)
(0.585,-0.4337)
(0.59,-0.4271)
(0.595,-0.4202)
(0.6,-0.4129)
(0.605,-0.4052)
(0.61,-0.3973)
(0.615,-0.389)
(0.62,-0.3805)
(0.625,-0.3717)
(0.63,-0.3626)
(0.635,-0.3533)
(0.64,-0.3438)
(0.645,-0.3341)
(0.65,-0.3242)
(0.655,-0.3141)
(0.66,-0.3039)
(0.665,-0.2936)
(0.67,-0.2831)
(0.675,-0.2726)
(0.68,-0.2619)
(0.685,-0.2512)
(0.69,-0.2405)
(0.695,-0.2297)
(0.7,-0.2189)
(0.705,-0.2082)
(0.71,-0.1974)
(0.715,-0.1867)
(0.72,-0.1761)
(0.725,-0.1655)
(0.73,-0.155)
(0.735,-0.1446)
(0.74,-0.1343)
(0.745,-0.1242)
(0.75,-0.1142)
(0.755,-0.1044)
(0.76,-0.0947)
(0.765,-0.0853)
(0.77,-0.076)
(0.775,-0.067)
(0.78,-0.0581)
(0.785,-0.0495)
(0.79,-0.0412)
(0.795,-0.0331)
(0.8,-0.0253)
(0.805,-0.0178)
(0.81,-0.0105)
(0.815,-0.0036)
(0.82,0.003)
(0.825,0.0094)
(0.83,0.0154)
(0.835,0.0211)
(0.84,0.0264)
(0.845,0.0314)
(0.85,0.0361)
(0.855,0.0404)
(0.86,0.0444)
(0.865,0.0481)
(0.87,0.0513)
(0.875,0.0542)
(0.88,0.0568)
(0.885,0.059)
(0.89,0.0609)
(0.895,0.0624)
(0.9,0.0635)
(0.905,0.0643)
(0.91,0.0647)
(0.915,0.0648)
(0.92,0.0645)
(0.925,0.0639)
(0.93,0.0629)
(0.935,0.0617)
(0.94,0.0601)
(0.945,0.0581)
(0.95,0.0559)
(0.955,0.0533)
(0.96,0.0504)
(0.965,0.0473)
(0.97,0.0439)
(0.975,0.0401)
(0.98,0.0361)
(0.985,0.0319)
(0.99,0.0274)
(0.995,0.0227)
(1.0,0.0177)
(1.005,0.0125)
(1.01,0.0071)
(1.015,0.0015)
(1.02,-0.0043)
(1.025,-0.0102)
(1.03,-0.0163)
(1.035,-0.0226)
(1.04,-0.029)
(1.045,-0.0355)
(1.05,-0.0422)
(1.055,-0.0489)
(1.06,-0.0557)
(1.065,-0.0626)
(1.07,-0.0696)
(1.075,-0.0766)
(1.08,-0.0837)
(1.085,-0.0907)
(1.09,-0.0978)
(1.095,-0.1049)
(1.1,-0.112)
(1.105,-0.119)
(1.11,-0.126)
(1.115,-0.1329)
(1.12,-0.1398)
(1.125,-0.1466)
(1.13,-0.1533)
(1.135,-0.1599)
(1.14,-0.1664)
(1.145,-0.1727)
(1.15,-0.179)
(1.155,-0.185)
(1.16,-0.1909)
(1.165,-0.1967)
(1.17,-0.2023)
(1.175,-0.2076)
(1.18,-0.2128)
(1.185,-0.2178)
(1.19,-0.2225)
(1.195,-0.2271)
(1.2,-0.2314)
(1.205,-0.2354)
(1.21,-0.2392)
(1.215,-0.2428)
(1.22,-0.2461)
(1.225,-0.2491)
(1.23,-0.2519)
(1.235,-0.2544)
(1.24,-0.2566)
(1.245,-0.2585)
(1.25,-0.2601)
(1.255,-0.2614)
(1.26,-0.2625)
(1.265,-0.2632)
(1.27,-0.2637)
(1.275,-0.2638)
(1.28,-0.2636)
(1.285,-0.2632)
(1.29,-0.2624)
(1.295,-0.2613)
(1.3,-0.26)
(1.305,-0.2583)
(1.31,-0.2564)
(1.315,-0.2541)
(1.32,-0.2516)
(1.325,-0.2487)
(1.33,-0.2456)
(1.335,-0.2422)
(1.34,-0.2385)
(1.345,-0.2346)
(1.35,-0.2304)
(1.355,-0.2259)
(1.36,-0.2212)
(1.365,-0.2162)
(1.37,-0.211)
(1.375,-0.2056)
(1.38,-0.1999)
(1.385,-0.1941)
(1.39,-0.188)
(1.395,-0.1817)
(1.4,-0.1752)
(1.405,-0.1686)
(1.41,-0.1618)
(1.415,-0.1548)
(1.42,-0.1476)
(1.425,-0.1403)
(1.43,-0.1329)
(1.435,-0.1254)
(1.44,-0.1178)
(1.445,-0.11)
(1.45,-0.1022)
(1.455,-0.0943)
(1.46,-0.0863)
(1.465,-0.0783)
(1.47,-0.0702)
(1.475,-0.0621)
(1.48,-0.0539)
(1.485,-0.0458)
(1.49,-0.0377)
(1.495,-0.0295)
(1.5,-0.0214)
(1.505,-0.0134)
(1.51,-0.0053)
(1.515,0.0026)
(1.52,0.0105)
(1.525,0.0184)
(1.53,0.0261)
(1.535,0.0337)
(1.54,0.0412)
(1.545,0.0486)
(1.55,0.0559)
(1.555,0.0631)
(1.56,0.07)
(1.565,0.0769)
(1.57,0.0835)
(1.575,0.09)
(1.58,0.0963)
(1.585,0.1024)
(1.59,0.1083)
(1.595,0.114)
(1.6,0.1195)
(1.605,0.1248)
(1.61,0.1298)
(1.615,0.1346)
(1.62,0.1392)
(1.625,0.1435)
(1.63,0.1475)
(1.635,0.1514)
(1.64,0.1549)
(1.645,0.1582)
(1.65,0.1612)
(1.655,0.164)
(1.66,0.1665)
(1.665,0.1687)
(1.67,0.1706)
(1.675,0.1723)
(1.68,0.1737)
(1.685,0.1748)
(1.69,0.1756)
(1.695,0.1762)
(1.7,0.1765)
(1.705,0.1765)
(1.71,0.1762)
(1.715,0.1757)
(1.72,0.1748)
(1.725,0.1738)
(1.73,0.1724)
(1.735,0.1708)
(1.74,0.169)
(1.745,0.1669)
(1.75,0.1645)
(1.755,0.162)
(1.76,0.1591)
(1.765,0.1561)
(1.77,0.1528)
(1.775,0.1493)
(1.78,0.1456)
(1.785,0.1417)
(1.79,0.1376)
(1.795,0.1333)
(1.8,0.1289)
(1.805,0.1242)
(1.81,0.1194)
(1.815,0.1145)
(1.82,0.1094)
(1.825,0.1041)
(1.83,0.0988)
(1.835,0.0933)
(1.84,0.0877)
(1.845,0.082)
(1.85,0.0762)
(1.855,0.0704)
(1.86,0.0644)
(1.865,0.0584)
(1.87,0.0524)
(1.875,0.0463)
(1.88,0.0402)
(1.885,0.0341)
(1.89,0.0279)
(1.895,0.0218)
(1.9,0.0157)
(1.905,0.0096)
(1.91,0.0035)
(1.915,-0.0025)
(1.92,-0.0085)
(1.925,-0.0144)
(1.93,-0.0202)
(1.935,-0.026)
(1.94,-0.0317)
(1.945,-0.0373)
(1.95,-0.0427)
(1.955,-0.0481)
(1.96,-0.0533)
(1.965,-0.0584)
(1.97,-0.0633)
(1.975,-0.0681)
(1.98,-0.0728)
(1.985,-0.0773)
(1.99,-0.0816)
(1.995,-0.0857)
(2.0,-0.0897)
(2.005,-0.0934)
(2.01,-0.097)
(2.015,-0.1003)
(2.02,-0.1035)
(2.025,-0.1064)
(2.03,-0.1091)
(2.035,-0.1116)
(2.04,-0.1139)
(2.045,-0.1159)
(2.05,-0.1178)
(2.055,-0.1193)
(2.06,-0.1207)
(2.065,-0.1218)
(2.07,-0.1226)
(2.075,-0.1233)
(2.08,-0.1237)
(2.085,-0.1238)
(2.09,-0.1237)
(2.095,-0.1234)
(2.1,-0.1228)
(2.105,-0.1219)
(2.11,-0.1209)
(2.115,-0.1196)
(2.12,-0.1181)
(2.125,-0.1163)
(2.13,-0.1143)
(2.135,-0.1121)
(2.14,-0.1097)
(2.145,-0.107)
(2.15,-0.1041)
(2.155,-0.1011)
(2.16,-0.0978)
(2.165,-0.0943)
(2.17,-0.0907)
(2.175,-0.0868)
(2.18,-0.0828)
(2.185,-0.0786)
(2.19,-0.0742)
(2.195,-0.0697)
(2.2,-0.065)
(2.205,-0.0602)
(2.21,-0.0552)
(2.215,-0.0502)
(2.22,-0.045)
(2.225,-0.0397)
(2.23,-0.0342)
(2.235,-0.0287)
(2.24,-0.0231)
(2.245,-0.0175)
(2.25,-0.0117)
(2.255,-0.006)
(2.26,-0.0001)
(2.265,0.0058)
(2.27,0.0117)
(2.275,0.0176)
(2.28,0.0235)
(2.285,0.0295)
(2.29,0.0354)
(2.295,0.0413)
(2.3,0.0471)
(2.305,0.053)
(2.31,0.0588)
(2.315,0.0645)
(2.32,0.0702)
(2.325,0.0757)
(2.33,0.0812)
(2.335,0.0866)
(2.34,0.0919)
(2.345,0.0971)
(2.35,0.1022)
(2.355,0.1071)
(2.36,0.1119)
(2.365,0.1166)
(2.37,0.1211)
(2.375,0.1254)
(2.38,0.1296)
(2.385,0.1336)
(2.39,0.1374)
(2.395,0.141)
(2.4,0.1445)
(2.405,0.1477)
(2.41,0.1507)
(2.415,0.1535)
(2.42,0.1561)
(2.425,0.1585)
(2.43,0.1606)
(2.435,0.1625)
(2.44,0.1642)
(2.445,0.1657)
(2.45,0.1669)
(2.455,0.1678)
(2.46,0.1685)
(2.465,0.169)
(2.47,0.1692)
(2.475,0.1692)
(2.48,0.1689)
(2.485,0.1683)
(2.49,0.1676)
(2.495,0.1665)
(2.5,0.1652)};
\addplot[name path=negative_top,color=red!70] coordinates {(0.005,1.0087)
(0.01,1.0075)
(0.015,1.0056)
(0.02,1.0031)
(0.025,1.0002)
(0.03,0.9969)
(0.035,0.9934)
(0.04,0.9897)
(0.045,0.9859)
(0.05,0.9819)
(0.055,0.9777)
(0.06,0.9733)
(0.065,0.9686)
(0.07,0.9636)
(0.075,0.9583)
(0.08,0.9527)
(0.085,0.9468)
(0.09,0.9406)
(0.095,0.9341)
(0.1,0.9273)
(0.105,0.9202)
(0.11,0.9127)
(0.115,0.905)
(0.12,0.8969)
(0.125,0.8886)
(0.13,0.8799)
(0.135,0.871)
(0.14,0.8617)
(0.145,0.8522)
(0.15,0.8424)
(0.155,0.8323)
(0.16,0.822)
(0.165,0.8114)
(0.17,0.8005)
(0.175,0.7893)
(0.18,0.778)
(0.185,0.7663)
(0.19,0.7545)
(0.195,0.7424)
(0.2,0.73)
(0.205,0.7175)
(0.21,0.7047)
(0.215,0.6918)
(0.22,0.6786)
(0.225,0.6652)
(0.23,0.6517)
(0.235,0.638)
(0.24,0.6241)
(0.245,0.61)
(0.25,0.5958)
(0.255,0.5814)
(0.26,0.5669)
(0.265,0.5523)
(0.27,0.5375)
(0.275,0.5226)
(0.28,0.5076)
(0.285,0.4925)
(0.29,0.4773)
(0.295,0.4621)
(0.3,0.4467)
(0.305,0.4313)
(0.31,0.4158)
(0.315,0.4003)
(0.32,0.3848)
(0.325,0.3692)
(0.33,0.3536)
(0.335,0.338)
(0.34,0.3223)
(0.345,0.3067)
(0.35,0.2911)
(0.355,0.2756)
(0.36,0.26)
(0.365,0.2446)
(0.37,0.2291)
(0.375,0.2138)
(0.38,0.1985)
(0.385,0.1833)
(0.39,0.1682)
(0.395,0.1532)
(0.4,0.1384)
(0.405,0.1237)
(0.41,0.1091)
(0.415,0.0947)
(0.42,0.0804)
(0.425,0.0664)
(0.43,0.0525)
(0.435,0.0389)
(0.44,0.0254)
(0.445,0.0122)
(0.45,-0.0007)
(0.455,-0.0134)
(0.46,-0.0258)
(0.465,-0.0378)
(0.47,-0.0496)
(0.475,-0.061)
(0.48,-0.0721)
(0.485,-0.0828)
(0.49,-0.093)
(0.495,-0.1029)
(0.5,-0.1122)
(0.505,-0.1211)
(0.51,-0.1295)
(0.515,-0.1373)
(0.52,-0.1446)
(0.525,-0.1512)
(0.53,-0.1572)
(0.535,-0.1624)
(0.54,-0.167)
(0.545,-0.1707)
(0.55,-0.1736)
(0.555,-0.1756)
(0.56,-0.1766)
(0.565,-0.1767)
(0.57,-0.1757)
(0.575,-0.1737)
(0.58,-0.1705)
(0.585,-0.1662)
(0.59,-0.1608)
(0.595,-0.1541)
(0.6,-0.1463)
(0.605,-0.1374)
(0.61,-0.1273)
(0.615,-0.1161)
(0.62,-0.1039)
(0.625,-0.0907)
(0.63,-0.0765)
(0.635,-0.0615)
(0.64,-0.0457)
(0.645,-0.0292)
(0.65,-0.012)
(0.655,0.0058)
(0.66,0.024)
(0.665,0.0426)
(0.67,0.0617)
(0.675,0.0809)
(0.68,0.1004)
(0.685,0.1201)
(0.69,0.1399)
(0.695,0.1597)
(0.7,0.1794)
(0.705,0.1992)
(0.71,0.2188)
(0.715,0.2382)
(0.72,0.2574)
(0.725,0.2764)
(0.73,0.2951)
(0.735,0.3134)
(0.74,0.3314)
(0.745,0.349)
(0.75,0.3661)
(0.755,0.3828)
(0.76,0.399)
(0.765,0.4147)
(0.77,0.4298)
(0.775,0.4444)
(0.78,0.4584)
(0.785,0.4717)
(0.79,0.4845)
(0.795,0.4965)
(0.8,0.5079)
(0.805,0.5187)
(0.81,0.5287)
(0.815,0.538)
(0.82,0.5467)
(0.825,0.5546)
(0.83,0.5617)
(0.835,0.5682)
(0.84,0.5739)
(0.845,0.5788)
(0.85,0.583)
(0.855,0.5865)
(0.86,0.5893)
(0.865,0.5913)
(0.87,0.5926)
(0.875,0.5932)
(0.88,0.593)
(0.885,0.5922)
(0.89,0.5907)
(0.895,0.5885)
(0.9,0.5857)
(0.905,0.5823)
(0.91,0.5782)
(0.915,0.5735)
(0.92,0.5683)
(0.925,0.5625)
(0.93,0.5562)
(0.935,0.5494)
(0.94,0.5421)
(0.945,0.5344)
(0.95,0.5263)
(0.955,0.5178)
(0.96,0.509)
(0.965,0.4998)
(0.97,0.4904)
(0.975,0.4808)
(0.98,0.471)
(0.985,0.461)
(0.99,0.4509)
(0.995,0.4408)
(1.0,0.4306)
(1.005,0.4204)
(1.01,0.4103)
(1.015,0.4002)
(1.02,0.3903)
(1.025,0.3805)
(1.03,0.3709)
(1.035,0.3615)
(1.04,0.3524)
(1.045,0.3435)
(1.05,0.335)
(1.055,0.3267)
(1.06,0.3188)
(1.065,0.3111)
(1.07,0.3039)
(1.075,0.2969)
(1.08,0.2904)
(1.085,0.2841)
(1.09,0.2782)
(1.095,0.2726)
(1.1,0.2673)
(1.105,0.2623)
(1.11,0.2576)
(1.115,0.2531)
(1.12,0.2489)
(1.125,0.2449)
(1.13,0.2412)
(1.135,0.2376)
(1.14,0.2342)
(1.145,0.2309)
(1.15,0.2278)
(1.155,0.2247)
(1.16,0.2218)
(1.165,0.219)
(1.17,0.2162)
(1.175,0.2135)
(1.18,0.2109)
(1.185,0.2083)
(1.19,0.2056)
(1.195,0.2031)
(1.2,0.2005)
(1.205,0.1979)
(1.21,0.1952)
(1.215,0.1926)
(1.22,0.1899)
(1.225,0.1873)
(1.23,0.1845)
(1.235,0.1818)
(1.24,0.179)
(1.245,0.1762)
(1.25,0.1733)
(1.255,0.1704)
(1.26,0.1675)
(1.265,0.1645)
(1.27,0.1615)
(1.275,0.1585)
(1.28,0.1555)
(1.285,0.1525)
(1.29,0.1494)
(1.295,0.1464)
(1.3,0.1434)
(1.305,0.1403)
(1.31,0.1374)
(1.315,0.1345)
(1.32,0.1316)
(1.325,0.1288)
(1.33,0.1261)
(1.335,0.1235)
(1.34,0.1211)
(1.345,0.1188)
(1.35,0.1166)
(1.355,0.1147)
(1.36,0.1129)
(1.365,0.1114)
(1.37,0.1102)
(1.375,0.1093)
(1.38,0.1087)
(1.385,0.1085)
(1.39,0.1087)
(1.395,0.1094)
(1.4,0.1105)
(1.405,0.1122)
(1.41,0.1144)
(1.415,0.1172)
(1.42,0.1207)
(1.425,0.1248)
(1.43,0.1297)
(1.435,0.1353)
(1.44,0.1416)
(1.445,0.1486)
(1.45,0.1564)
(1.455,0.165)
(1.46,0.1743)
(1.465,0.1842)
(1.47,0.1949)
(1.475,0.2062)
(1.48,0.218)
(1.485,0.2304)
(1.49,0.2433)
(1.495,0.2566)
(1.5,0.2702)
(1.505,0.2842)
(1.51,0.2984)
(1.515,0.3127)
(1.52,0.3273)
(1.525,0.3418)
(1.53,0.3564)
(1.535,0.371)
(1.54,0.3855)
(1.545,0.3998)
(1.55,0.414)
(1.555,0.428)
(1.56,0.4417)
(1.565,0.4551)
(1.57,0.4682)
(1.575,0.4809)
(1.58,0.4932)
(1.585,0.5051)
(1.59,0.5166)
(1.595,0.5276)
(1.6,0.538)
(1.605,0.548)
(1.61,0.5574)
(1.615,0.5662)
(1.62,0.5744)
(1.625,0.582)
(1.63,0.589)
(1.635,0.5954)
(1.64,0.6011)
(1.645,0.6062)
(1.65,0.6106)
(1.655,0.6144)
(1.66,0.6174)
(1.665,0.6198)
(1.67,0.6215)
(1.675,0.6225)
(1.68,0.6228)
(1.685,0.6225)
(1.69,0.6214)
(1.695,0.6197)
(1.7,0.6173)
(1.705,0.6142)
(1.71,0.6104)
(1.715,0.606)
(1.72,0.601)
(1.725,0.5953)
(1.73,0.5891)
(1.735,0.5822)
(1.74,0.5747)
(1.745,0.5667)
(1.75,0.5581)
(1.755,0.549)
(1.76,0.5394)
(1.765,0.5293)
(1.77,0.5188)
(1.775,0.5079)
(1.78,0.4966)
(1.785,0.4849)
(1.79,0.473)
(1.795,0.4607)
(1.8,0.4483)
(1.805,0.4356)
(1.81,0.4228)
(1.815,0.41)
(1.82,0.3971)
(1.825,0.3842)
(1.83,0.3714)
(1.835,0.3588)
(1.84,0.3464)
(1.845,0.3343)
(1.85,0.3226)
(1.855,0.3112)
(1.86,0.3005)
(1.865,0.2902)
(1.87,0.2807)
(1.875,0.2718)
(1.88,0.2637)
(1.885,0.2563)
(1.89,0.2498)
(1.895,0.2442)
(1.9,0.2393)
(1.905,0.2353)
(1.91,0.232)
(1.915,0.2296)
(1.92,0.2278)
(1.925,0.2267)
(1.93,0.2262)
(1.935,0.2263)
(1.94,0.2269)
(1.945,0.228)
(1.95,0.2294)
(1.955,0.2312)
(1.96,0.2334)
(1.965,0.2358)
(1.97,0.2384)
(1.975,0.2412)
(1.98,0.2441)
(1.985,0.2472)
(1.99,0.2504)
(1.995,0.2537)
(2.0,0.257)
(2.005,0.2604)
(2.01,0.2638)
(2.015,0.2671)
(2.02,0.2704)
(2.025,0.2737)
(2.03,0.277)
(2.035,0.2802)
(2.04,0.2833)
(2.045,0.2863)
(2.05,0.2892)
(2.055,0.292)
(2.06,0.2947)
(2.065,0.2973)
(2.07,0.2997)
(2.075,0.3021)
(2.08,0.3043)
(2.085,0.3063)
(2.09,0.3082)
(2.095,0.31)
(2.1,0.3116)
(2.105,0.3131)
(2.11,0.3144)
(2.115,0.3155)
(2.12,0.3165)
(2.125,0.3174)
(2.13,0.318)
(2.135,0.3186)
(2.14,0.319)
(2.145,0.3192)
(2.15,0.3193)
(2.155,0.3192)
(2.16,0.319)
(2.165,0.3187)
(2.17,0.3182)
(2.175,0.3176)
(2.18,0.3169)
(2.185,0.3161)
(2.19,0.3152)
(2.195,0.3142)
(2.2,0.3131)
(2.205,0.312)
(2.21,0.3107)
(2.215,0.3095)
(2.22,0.3081)
(2.225,0.3068)
(2.23,0.3055)
(2.235,0.3042)
(2.24,0.3029)
(2.245,0.3017)
(2.25,0.3005)
(2.255,0.2994)
(2.26,0.2985)
(2.265,0.2977)
(2.27,0.2972)
(2.275,0.2968)
(2.28,0.2966)
(2.285,0.2968)
(2.29,0.2972)
(2.295,0.298)
(2.3,0.2991)
(2.305,0.3007)
(2.31,0.3027)
(2.315,0.3051)
(2.32,0.308)
(2.325,0.3115)
(2.33,0.3154)
(2.335,0.3198)
(2.34,0.3247)
(2.345,0.3301)
(2.35,0.336)
(2.355,0.3423)
(2.36,0.349)
(2.365,0.3561)
(2.37,0.3636)
(2.375,0.3713)
(2.38,0.3792)
(2.385,0.3873)
(2.39,0.3956)
(2.395,0.4039)
(2.4,0.4123)
(2.405,0.4206)
(2.41,0.4289)
(2.415,0.4371)
(2.42,0.4451)
(2.425,0.453)
(2.43,0.4606)
(2.435,0.468)
(2.44,0.475)
(2.445,0.4818)
(2.45,0.4882)
(2.455,0.4942)
(2.46,0.4999)
(2.465,0.5051)
(2.47,0.5099)
(2.475,0.5142)
(2.48,0.5181)
(2.485,0.5215)
(2.49,0.5244)
(2.495,0.5268)
(2.5,0.5287)};
\addplot[name path=negative_down,color=red!70] coordinates {(0.005,0.9872)
(0.01,0.9861)
(0.015,0.9843)
(0.02,0.9816)
(0.025,0.9779)
(0.03,0.973)
(0.035,0.9669)
(0.04,0.9594)
(0.045,0.9507)
(0.05,0.9408)
(0.055,0.9296)
(0.06,0.9173)
(0.065,0.9039)
(0.07,0.8894)
(0.075,0.8738)
(0.08,0.8573)
(0.085,0.8397)
(0.09,0.8212)
(0.095,0.8018)
(0.1,0.7815)
(0.105,0.7603)
(0.11,0.7383)
(0.115,0.7154)
(0.12,0.6918)
(0.125,0.6674)
(0.13,0.6423)
(0.135,0.6165)
(0.14,0.5901)
(0.145,0.563)
(0.15,0.5353)
(0.155,0.5071)
(0.16,0.4784)
(0.165,0.4492)
(0.17,0.4195)
(0.175,0.3895)
(0.18,0.359)
(0.185,0.3283)
(0.19,0.2972)
(0.195,0.2659)
(0.2,0.2344)
(0.205,0.2027)
(0.21,0.1708)
(0.215,0.1389)
(0.22,0.1069)
(0.225,0.0748)
(0.23,0.0428)
(0.235,0.0109)
(0.24,-0.021)
(0.245,-0.0527)
(0.25,-0.0843)
(0.255,-0.1156)
(0.26,-0.1467)
(0.265,-0.1776)
(0.27,-0.2081)
(0.275,-0.2382)
(0.28,-0.268)
(0.285,-0.2973)
(0.29,-0.3262)
(0.295,-0.3546)
(0.3,-0.3825)
(0.305,-0.4098)
(0.31,-0.4365)
(0.315,-0.4626)
(0.32,-0.4881)
(0.325,-0.5129)
(0.33,-0.537)
(0.335,-0.5604)
(0.34,-0.583)
(0.345,-0.6049)
(0.35,-0.6259)
(0.355,-0.6462)
(0.36,-0.6656)
(0.365,-0.6842)
(0.37,-0.7019)
(0.375,-0.7187)
(0.38,-0.7346)
(0.385,-0.7497)
(0.39,-0.7638)
(0.395,-0.7769)
(0.4,-0.7892)
(0.405,-0.8005)
(0.41,-0.8108)
(0.415,-0.8202)
(0.42,-0.8287)
(0.425,-0.8362)
(0.43,-0.8428)
(0.435,-0.8484)
(0.44,-0.8531)
(0.445,-0.8569)
(0.45,-0.8597)
(0.455,-0.8617)
(0.46,-0.8627)
(0.465,-0.8629)
(0.47,-0.8623)
(0.475,-0.8608)
(0.48,-0.8585)
(0.485,-0.8555)
(0.49,-0.8517)
(0.495,-0.8472)
(0.5,-0.842)
(0.505,-0.8361)
(0.51,-0.8297)
(0.515,-0.8227)
(0.52,-0.8153)
(0.525,-0.8074)
(0.53,-0.7991)
(0.535,-0.7904)
(0.54,-0.7815)
(0.545,-0.7724)
(0.55,-0.7632)
(0.555,-0.754)
(0.56,-0.7447)
(0.565,-0.7355)
(0.57,-0.7266)
(0.575,-0.7178)
(0.58,-0.7093)
(0.585,-0.7012)
(0.59,-0.6935)
(0.595,-0.6862)
(0.6,-0.6794)
(0.605,-0.6731)
(0.61,-0.6673)
(0.615,-0.6619)
(0.62,-0.6571)
(0.625,-0.6527)
(0.63,-0.6487)
(0.635,-0.6452)
(0.64,-0.6419)
(0.645,-0.639)
(0.65,-0.6364)
(0.655,-0.634)
(0.66,-0.6318)
(0.665,-0.6298)
(0.67,-0.6279)
(0.675,-0.6261)
(0.68,-0.6243)
(0.685,-0.6226)
(0.69,-0.6209)
(0.695,-0.6191)
(0.7,-0.6173)
(0.705,-0.6155)
(0.71,-0.6136)
(0.715,-0.6116)
(0.72,-0.6096)
(0.725,-0.6074)
(0.73,-0.6051)
(0.735,-0.6026)
(0.74,-0.6001)
(0.745,-0.5974)
(0.75,-0.5946)
(0.755,-0.5916)
(0.76,-0.5885)
(0.765,-0.5852)
(0.77,-0.5819)
(0.775,-0.5783)
(0.78,-0.5746)
(0.785,-0.5708)
(0.79,-0.5669)
(0.795,-0.5628)
(0.8,-0.5586)
(0.805,-0.5543)
(0.81,-0.5498)
(0.815,-0.5452)
(0.82,-0.5406)
(0.825,-0.5358)
(0.83,-0.531)
(0.835,-0.526)
(0.84,-0.521)
(0.845,-0.516)
(0.85,-0.5108)
(0.855,-0.5056)
(0.86,-0.5004)
(0.865,-0.4952)
(0.87,-0.4899)
(0.875,-0.4847)
(0.88,-0.4794)
(0.885,-0.4742)
(0.89,-0.469)
(0.895,-0.4638)
(0.9,-0.4587)
(0.905,-0.4537)
(0.91,-0.4488)
(0.915,-0.444)
(0.92,-0.4393)
(0.925,-0.4347)
(0.93,-0.4303)
(0.935,-0.4261)
(0.94,-0.422)
(0.945,-0.4182)
(0.95,-0.4145)
(0.955,-0.4112)
(0.96,-0.4081)
(0.965,-0.4052)
(0.97,-0.4027)
(0.975,-0.4005)
(0.98,-0.3987)
(0.985,-0.3972)
(0.99,-0.3961)
(0.995,-0.3955)
(1.0,-0.3952)
(1.005,-0.3954)
(1.01,-0.3961)
(1.015,-0.3972)
(1.02,-0.3988)
(1.025,-0.4009)
(1.03,-0.4036)
(1.035,-0.4067)
(1.04,-0.4104)
(1.045,-0.4146)
(1.05,-0.4193)
(1.055,-0.4245)
(1.06,-0.4302)
(1.065,-0.4364)
(1.07,-0.4431)
(1.075,-0.4502)
(1.08,-0.4577)
(1.085,-0.4656)
(1.09,-0.4738)
(1.095,-0.4824)
(1.1,-0.4912)
(1.105,-0.5003)
(1.11,-0.5095)
(1.115,-0.519)
(1.12,-0.5285)
(1.125,-0.5381)
(1.13,-0.5477)
(1.135,-0.5573)
(1.14,-0.5669)
(1.145,-0.5764)
(1.15,-0.5857)
(1.155,-0.5948)
(1.16,-0.6037)
(1.165,-0.6124)
(1.17,-0.6208)
(1.175,-0.6288)
(1.18,-0.6365)
(1.185,-0.6438)
(1.19,-0.6507)
(1.195,-0.6572)
(1.2,-0.6632)
(1.205,-0.6687)
(1.21,-0.6737)
(1.215,-0.6782)
(1.22,-0.6821)
(1.225,-0.6855)
(1.23,-0.6883)
(1.235,-0.6905)
(1.24,-0.6921)
(1.245,-0.6931)
(1.25,-0.6935)
(1.255,-0.6933)
(1.26,-0.6924)
(1.265,-0.6909)
(1.27,-0.6888)
(1.275,-0.6861)
(1.28,-0.6828)
(1.285,-0.6788)
(1.29,-0.6742)
(1.295,-0.6691)
(1.3,-0.6633)
(1.305,-0.657)
(1.31,-0.6501)
(1.315,-0.6427)
(1.32,-0.6347)
(1.325,-0.6263)
(1.33,-0.6173)
(1.335,-0.6079)
(1.34,-0.5981)
(1.345,-0.5879)
(1.35,-0.5774)
(1.355,-0.5665)
(1.36,-0.5553)
(1.365,-0.5439)
(1.37,-0.5323)
(1.375,-0.5205)
(1.38,-0.5086)
(1.385,-0.4966)
(1.39,-0.4847)
(1.395,-0.4728)
(1.4,-0.461)
(1.405,-0.4493)
(1.41,-0.4379)
(1.415,-0.4268)
(1.42,-0.416)
(1.425,-0.4055)
(1.43,-0.3956)
(1.435,-0.3861)
(1.44,-0.3771)
(1.445,-0.3687)
(1.45,-0.3608)
(1.455,-0.3535)
(1.46,-0.3468)
(1.465,-0.3408)
(1.47,-0.3353)
(1.475,-0.3303)
(1.48,-0.3259)
(1.485,-0.322)
(1.49,-0.3186)
(1.495,-0.3156)
(1.5,-0.3131)
(1.505,-0.3109)
(1.51,-0.309)
(1.515,-0.3075)
(1.52,-0.3062)
(1.525,-0.3051)
(1.53,-0.3043)
(1.535,-0.3036)
(1.54,-0.303)
(1.545,-0.3025)
(1.55,-0.3022)
(1.555,-0.3019)
(1.56,-0.3016)
(1.565,-0.3014)
(1.57,-0.3012)
(1.575,-0.3009)
(1.58,-0.3006)
(1.585,-0.3003)
(1.59,-0.3)
(1.595,-0.2995)
(1.6,-0.299)
(1.605,-0.2984)
(1.61,-0.2978)
(1.615,-0.297)
(1.62,-0.2961)
(1.625,-0.2951)
(1.63,-0.2939)
(1.635,-0.2927)
(1.64,-0.2913)
(1.645,-0.2898)
(1.65,-0.2882)
(1.655,-0.2864)
(1.66,-0.2845)
(1.665,-0.2824)
(1.67,-0.2803)
(1.675,-0.2779)
(1.68,-0.2755)
(1.685,-0.2729)
(1.69,-0.2702)
(1.695,-0.2673)
(1.7,-0.2644)
(1.705,-0.2613)
(1.71,-0.2581)
(1.715,-0.2547)
(1.72,-0.2513)
(1.725,-0.2478)
(1.73,-0.2442)
(1.735,-0.2405)
(1.74,-0.2367)
(1.745,-0.2329)
(1.75,-0.229)
(1.755,-0.2251)
(1.76,-0.2211)
(1.765,-0.2172)
(1.77,-0.2132)
(1.775,-0.2093)
(1.78,-0.2054)
(1.785,-0.2015)
(1.79,-0.1977)
(1.795,-0.1941)
(1.8,-0.1905)
(1.805,-0.1872)
(1.81,-0.184)
(1.815,-0.181)
(1.82,-0.1783)
(1.825,-0.1759)
(1.83,-0.1739)
(1.835,-0.1722)
(1.84,-0.171)
(1.845,-0.1703)
(1.85,-0.1701)
(1.855,-0.1705)
(1.86,-0.1716)
(1.865,-0.1734)
(1.87,-0.1759)
(1.875,-0.1792)
(1.88,-0.1833)
(1.885,-0.1882)
(1.89,-0.194)
(1.895,-0.2005)
(1.9,-0.2079)
(1.905,-0.2161)
(1.91,-0.225)
(1.915,-0.2346)
(1.92,-0.2448)
(1.925,-0.2555)
(1.93,-0.2667)
(1.935,-0.2783)
(1.94,-0.2903)
(1.945,-0.3025)
(1.95,-0.3149)
(1.955,-0.3274)
(1.96,-0.34)
(1.965,-0.3525)
(1.97,-0.3651)
(1.975,-0.3775)
(1.98,-0.3897)
(1.985,-0.4018)
(1.99,-0.4136)
(1.995,-0.4251)
(2.0,-0.4363)
(2.005,-0.4472)
(2.01,-0.4577)
(2.015,-0.4677)
(2.02,-0.4774)
(2.025,-0.4865)
(2.03,-0.4952)
(2.035,-0.5034)
(2.04,-0.511)
(2.045,-0.5181)
(2.05,-0.5247)
(2.055,-0.5307)
(2.06,-0.5361)
(2.065,-0.5408)
(2.07,-0.545)
(2.075,-0.5486)
(2.08,-0.5516)
(2.085,-0.5539)
(2.09,-0.5556)
(2.095,-0.5567)
(2.1,-0.5571)
(2.105,-0.557)
(2.11,-0.5561)
(2.115,-0.5547)
(2.12,-0.5526)
(2.125,-0.55)
(2.13,-0.5467)
(2.135,-0.5428)
(2.14,-0.5383)
(2.145,-0.5332)
(2.15,-0.5276)
(2.155,-0.5214)
(2.16,-0.5146)
(2.165,-0.5073)
(2.17,-0.4996)
(2.175,-0.4913)
(2.18,-0.4825)
(2.185,-0.4733)
(2.19,-0.4637)
(2.195,-0.4536)
(2.2,-0.4432)
(2.205,-0.4324)
(2.21,-0.4212)
(2.215,-0.4098)
(2.22,-0.3981)
(2.225,-0.3861)
(2.23,-0.374)
(2.235,-0.3616)
(2.24,-0.3492)
(2.245,-0.3366)
(2.25,-0.324)
(2.255,-0.3113)
(2.26,-0.2987)
(2.265,-0.2862)
(2.27,-0.2738)
(2.275,-0.2616)
(2.28,-0.2496)
(2.285,-0.2379)
(2.29,-0.2265)
(2.295,-0.2154)
(2.3,-0.2048)
(2.305,-0.1947)
(2.31,-0.1851)
(2.315,-0.1761)
(2.32,-0.1677)
(2.325,-0.16)
(2.33,-0.1529)
(2.335,-0.1465)
(2.34,-0.1408)
(2.345,-0.1359)
(2.35,-0.1316)
(2.355,-0.1281)
(2.36,-0.1252)
(2.365,-0.123)
(2.37,-0.1214)
(2.375,-0.1204)
(2.38,-0.12)
(2.385,-0.1202)
(2.39,-0.1208)
(2.395,-0.1219)
(2.4,-0.1234)
(2.405,-0.1253)
(2.41,-0.1275)
(2.415,-0.1301)
(2.42,-0.1329)
(2.425,-0.136)
(2.43,-0.1393)
(2.435,-0.1429)
(2.44,-0.1466)
(2.445,-0.1505)
(2.45,-0.1545)
(2.455,-0.1586)
(2.46,-0.1628)
(2.465,-0.1671)
(2.47,-0.1715)
(2.475,-0.1759)
(2.48,-0.1803)
(2.485,-0.1848)
(2.49,-0.1893)
(2.495,-0.1938)
(2.5,-0.1982)};
\addplot[red!50,fill opacity=0.5] fill between[of=negative_top and negative_down];

\addplot[color=blue] coordinates {(0.005,0.9953)
(0.01,0.9945)
(0.015,0.9932)
(0.02,0.9914)
(0.025,0.9891)
(0.03,0.9863)
(0.035,0.983)
(0.04,0.9792)
(0.045,0.9748)
(0.05,0.97)
(0.055,0.9647)
(0.06,0.9589)
(0.065,0.9526)
(0.07,0.9459)
(0.075,0.9386)
(0.08,0.9309)
(0.085,0.9228)
(0.09,0.9142)
(0.095,0.9051)
(0.1,0.8956)
(0.105,0.8857)
(0.11,0.8753)
(0.115,0.8645)
(0.12,0.8534)
(0.125,0.8418)
(0.13,0.8298)
(0.135,0.8175)
(0.14,0.8048)
(0.145,0.7917)
(0.15,0.7783)
(0.155,0.7646)
(0.16,0.7505)
(0.165,0.7361)
(0.17,0.7214)
(0.175,0.7065)
(0.18,0.6912)
(0.185,0.6757)
(0.19,0.6599)
(0.195,0.6439)
(0.2,0.6276)
(0.205,0.6112)
(0.21,0.5945)
(0.215,0.5776)
(0.22,0.5606)
(0.225,0.5434)
(0.23,0.526)
(0.235,0.5085)
(0.24,0.4909)
(0.245,0.4731)
(0.25,0.4553)
(0.255,0.4373)
(0.26,0.4193)
(0.265,0.4013)
(0.27,0.3832)
(0.275,0.365)
(0.28,0.3469)
(0.285,0.3287)
(0.29,0.3105)
(0.295,0.2924)
(0.3,0.2742)
(0.305,0.2561)
(0.31,0.2381)
(0.315,0.2202)
(0.32,0.2023)
(0.325,0.1845)
(0.33,0.1668)
(0.335,0.1492)
(0.34,0.1318)
(0.345,0.1145)
(0.35,0.0973)
(0.355,0.0803)
(0.36,0.0635)
(0.365,0.0468)
(0.37,0.0304)
(0.375,0.0141)
(0.38,-0.0019)
(0.385,-0.0178)
(0.39,-0.0334)
(0.395,-0.0488)
(0.4,-0.0639)
(0.405,-0.0788)
(0.41,-0.0934)
(0.415,-0.1078)
(0.42,-0.1219)
(0.425,-0.1357)
(0.43,-0.1492)
(0.435,-0.1624)
(0.44,-0.1754)
(0.445,-0.188)
(0.45,-0.2003)
(0.455,-0.2123)
(0.46,-0.224)
(0.465,-0.2354)
(0.47,-0.2465)
(0.475,-0.2572)
(0.48,-0.2676)
(0.485,-0.2776)
(0.49,-0.2874)
(0.495,-0.2968)
(0.5,-0.3058)
(0.505,-0.3145)
(0.51,-0.3229)
(0.515,-0.3309)
(0.52,-0.3386)
(0.525,-0.346)
(0.53,-0.353)
(0.535,-0.3597)
(0.54,-0.366)
(0.545,-0.372)
(0.55,-0.3777)
(0.555,-0.383)
(0.56,-0.388)
(0.565,-0.3927)
(0.57,-0.397)
(0.575,-0.4011)
(0.58,-0.4048)
(0.585,-0.4082)
(0.59,-0.4113)
(0.595,-0.4141)
(0.6,-0.4165)
(0.605,-0.4187)
(0.61,-0.4206)
(0.615,-0.4222)
(0.62,-0.4236)
(0.625,-0.4246)
(0.63,-0.4254)
(0.635,-0.4259)
(0.64,-0.4262)
(0.645,-0.4262)
(0.65,-0.426)
(0.655,-0.4255)
(0.66,-0.4248)
(0.665,-0.4239)
(0.67,-0.4227)
(0.675,-0.4214)
(0.68,-0.4198)
(0.685,-0.4181)
(0.69,-0.4161)
(0.695,-0.414)
(0.7,-0.4117)
(0.705,-0.4092)
(0.71,-0.4066)
(0.715,-0.4038)
(0.72,-0.4009)
(0.725,-0.3979)
(0.73,-0.3947)
(0.735,-0.3914)
(0.74,-0.388)
(0.745,-0.3845)
(0.75,-0.3808)
(0.755,-0.3771)
(0.76,-0.3733)
(0.765,-0.3694)
(0.77,-0.3655)
(0.775,-0.3615)
(0.78,-0.3574)
(0.785,-0.3533)
(0.79,-0.3492)
(0.795,-0.345)
(0.8,-0.3408)
(0.805,-0.3365)
(0.81,-0.3323)
(0.815,-0.328)
(0.82,-0.3237)
(0.825,-0.3195)
(0.83,-0.3152)
(0.835,-0.311)
(0.84,-0.3067)
(0.845,-0.3025)
(0.85,-0.2984)
(0.855,-0.2942)
(0.86,-0.2901)
(0.865,-0.2861)
(0.87,-0.282)
(0.875,-0.2781)
(0.88,-0.2742)
(0.885,-0.2703)
(0.89,-0.2665)
(0.895,-0.2628)
(0.9,-0.2592)
(0.905,-0.2556)
(0.91,-0.2521)
(0.915,-0.2486)
(0.92,-0.2453)
(0.925,-0.242)
(0.93,-0.2388)
(0.935,-0.2357)
(0.94,-0.2327)
(0.945,-0.2298)
(0.95,-0.2269)
(0.955,-0.2242)
(0.96,-0.2215)
(0.965,-0.2189)
(0.97,-0.2164)
(0.975,-0.214)
(0.98,-0.2117)
(0.985,-0.2095)
(0.99,-0.2073)
(0.995,-0.2053)
(1.0,-0.2033)
(1.005,-0.2015)
(1.01,-0.1997)
(1.015,-0.198)
(1.02,-0.1963)
(1.025,-0.1948)
(1.03,-0.1933)
(1.035,-0.192)
(1.04,-0.1906)
(1.045,-0.1894)
(1.05,-0.1882)
(1.055,-0.1871)
(1.06,-0.1861)
(1.065,-0.1851)
(1.07,-0.1842)
(1.075,-0.1834)
(1.08,-0.1826)
(1.085,-0.1818)
(1.09,-0.1811)
(1.095,-0.1804)
(1.1,-0.1798)
(1.105,-0.1793)
(1.11,-0.1787)
(1.115,-0.1782)
(1.12,-0.1777)
(1.125,-0.1773)
(1.13,-0.1768)
(1.135,-0.1764)
(1.14,-0.176)
(1.145,-0.1756)
(1.15,-0.1752)
(1.155,-0.1749)
(1.16,-0.1745)
(1.165,-0.1741)
(1.17,-0.1737)
(1.175,-0.1733)
(1.18,-0.1729)
(1.185,-0.1725)
(1.19,-0.1721)
(1.195,-0.1716)
(1.2,-0.1712)
(1.205,-0.1706)
(1.21,-0.1701)
(1.215,-0.1695)
(1.22,-0.1689)
(1.225,-0.1683)
(1.23,-0.1676)
(1.235,-0.1669)
(1.24,-0.1661)
(1.245,-0.1653)
(1.25,-0.1644)
(1.255,-0.1635)
(1.26,-0.1625)
(1.265,-0.1615)
(1.27,-0.1604)
(1.275,-0.1592)
(1.28,-0.158)
(1.285,-0.1567)
(1.29,-0.1554)
(1.295,-0.154)
(1.3,-0.1525)
(1.305,-0.151)
(1.31,-0.1494)
(1.315,-0.1477)
(1.32,-0.146)
(1.325,-0.1441)
(1.33,-0.1423)
(1.335,-0.1403)
(1.34,-0.1383)
(1.345,-0.1362)
(1.35,-0.1341)
(1.355,-0.1318)
(1.36,-0.1296)
(1.365,-0.1272)
(1.37,-0.1248)
(1.375,-0.1223)
(1.38,-0.1197)
(1.385,-0.1171)
(1.39,-0.1145)
(1.395,-0.1117)
(1.4,-0.1089)
(1.405,-0.1061)
(1.41,-0.1032)
(1.415,-0.1002)
(1.42,-0.0972)
(1.425,-0.0941)
(1.43,-0.091)
(1.435,-0.0879)
(1.44,-0.0847)
(1.445,-0.0814)
(1.45,-0.0782)
(1.455,-0.0748)
(1.46,-0.0715)
(1.465,-0.0681)
(1.47,-0.0647)
(1.475,-0.0613)
(1.48,-0.0578)
(1.485,-0.0543)
(1.49,-0.0508)
(1.495,-0.0473)
(1.5,-0.0438)
(1.505,-0.0402)
(1.51,-0.0367)
(1.515,-0.0331)
(1.52,-0.0296)
(1.525,-0.0261)
(1.53,-0.0225)
(1.535,-0.019)
(1.54,-0.0155)
(1.545,-0.012)
(1.55,-0.0085)
(1.555,-0.005)
(1.56,-0.0016)
(1.565,0.0019)
(1.57,0.0053)
(1.575,0.0086)
(1.58,0.012)
(1.585,0.0152)
(1.59,0.0185)
(1.595,0.0217)
(1.6,0.0249)
(1.605,0.028)
(1.61,0.0311)
(1.615,0.0341)
(1.62,0.0371)
(1.625,0.04)
(1.63,0.0428)
(1.635,0.0456)
(1.64,0.0484)
(1.645,0.051)
(1.65,0.0536)
(1.655,0.0562)
(1.66,0.0586)
(1.665,0.061)
(1.67,0.0634)
(1.675,0.0656)
(1.68,0.0678)
(1.685,0.0699)
(1.69,0.072)
(1.695,0.0739)
(1.7,0.0758)
(1.705,0.0776)
(1.71,0.0793)
(1.715,0.081)
(1.72,0.0825)
(1.725,0.084)
(1.73,0.0854)
(1.735,0.0867)
(1.74,0.088)
(1.745,0.0891)
(1.75,0.0902)
(1.755,0.0912)
(1.76,0.0921)
(1.765,0.093)
(1.77,0.0937)
(1.775,0.0944)
(1.78,0.095)
(1.785,0.0956)
(1.79,0.096)
(1.795,0.0964)
(1.8,0.0968)
(1.805,0.097)
(1.81,0.0972)
(1.815,0.0973)
(1.82,0.0973)
(1.825,0.0973)
(1.83,0.0972)
(1.835,0.097)
(1.84,0.0968)
(1.845,0.0965)
(1.85,0.0962)
(1.855,0.0958)
(1.86,0.0954)
(1.865,0.0949)
(1.87,0.0943)
(1.875,0.0937)
(1.88,0.0931)
(1.885,0.0924)
(1.89,0.0917)
(1.895,0.0909)
(1.9,0.0901)
(1.905,0.0893)
(1.91,0.0884)
(1.915,0.0875)
(1.92,0.0866)
(1.925,0.0856)
(1.93,0.0846)
(1.935,0.0836)
(1.94,0.0826)
(1.945,0.0815)
(1.95,0.0805)
(1.955,0.0794)
(1.96,0.0783)
(1.965,0.0772)
(1.97,0.0761)
(1.975,0.075)
(1.98,0.0739)
(1.985,0.0727)
(1.99,0.0716)
(1.995,0.0705)
(2.0,0.0694)
(2.005,0.0683)
(2.01,0.0672)
(2.015,0.0661)
(2.02,0.065)
(2.025,0.0639)
(2.03,0.0628)
(2.035,0.0618)
(2.04,0.0607)
(2.045,0.0597)
(2.05,0.0587)
(2.055,0.0577)
(2.06,0.0568)
(2.065,0.0558)
(2.07,0.0549)
(2.075,0.054)
(2.08,0.0531)
(2.085,0.0523)
(2.09,0.0514)
(2.095,0.0506)
(2.1,0.0499)
(2.105,0.0491)
(2.11,0.0484)
(2.115,0.0477)
(2.12,0.047)
(2.125,0.0464)
(2.13,0.0457)
(2.135,0.0452)
(2.14,0.0446)
(2.145,0.044)
(2.15,0.0435)
(2.155,0.043)
(2.16,0.0426)
(2.165,0.0421)
(2.17,0.0417)
(2.175,0.0413)
(2.18,0.0409)
(2.185,0.0406)
(2.19,0.0403)
(2.195,0.0399)
(2.2,0.0396)
(2.205,0.0394)
(2.21,0.0391)
(2.215,0.0389)
(2.22,0.0387)
(2.225,0.0384)
(2.23,0.0382)
(2.235,0.0381)
(2.24,0.0379)
(2.245,0.0377)
(2.25,0.0376)
(2.255,0.0374)
(2.26,0.0373)
(2.265,0.0371)
(2.27,0.037)
(2.275,0.0368)
(2.28,0.0367)
(2.285,0.0365)
(2.29,0.0364)
(2.295,0.0362)
(2.3,0.0361)
(2.305,0.0359)
(2.31,0.0358)
(2.315,0.0356)
(2.32,0.0354)
(2.325,0.0352)
(2.33,0.0349)
(2.335,0.0347)
(2.34,0.0345)
(2.345,0.0342)
(2.35,0.0339)
(2.355,0.0336)
(2.36,0.0333)
(2.365,0.0329)
(2.37,0.0325)
(2.375,0.0321)
(2.38,0.0317)
(2.385,0.0312)
(2.39,0.0308)
(2.395,0.0303)
(2.4,0.0297)
(2.405,0.0292)
(2.41,0.0286)
(2.415,0.0279)
(2.42,0.0273)
(2.425,0.0266)
(2.43,0.0259)
(2.435,0.0251)
(2.44,0.0243)
(2.445,0.0235)
(2.45,0.0227)
(2.455,0.0218)
(2.46,0.0208)
(2.465,0.0199)
(2.47,0.0189)
(2.475,0.0179)
(2.48,0.0169)
(2.485,0.0158)
(2.49,0.0147)
(2.495,0.0135)
(2.5,0.0124)};
\addplot[name path=positive_top,color=blue!70] coordinates {(0.005,1.0133)
(0.01,1.0125)
(0.015,1.0112)
(0.02,1.0093)
(0.025,1.007)
(0.03,1.0042)
(0.035,1.001)
(0.04,0.9974)
(0.045,0.9935)
(0.05,0.9894)
(0.055,0.985)
(0.06,0.9804)
(0.065,0.9756)
(0.07,0.9707)
(0.075,0.9655)
(0.08,0.9602)
(0.085,0.9547)
(0.09,0.9489)
(0.095,0.943)
(0.1,0.9369)
(0.105,0.9305)
(0.11,0.9239)
(0.115,0.917)
(0.12,0.91)
(0.125,0.9027)
(0.13,0.8951)
(0.135,0.8874)
(0.14,0.8793)
(0.145,0.8711)
(0.15,0.8627)
(0.155,0.854)
(0.16,0.8451)
(0.165,0.8359)
(0.17,0.8266)
(0.175,0.817)
(0.18,0.8073)
(0.185,0.7973)
(0.19,0.7871)
(0.195,0.7767)
(0.2,0.7662)
(0.205,0.7554)
(0.21,0.7445)
(0.215,0.7334)
(0.22,0.7221)
(0.225,0.7107)
(0.23,0.6991)
(0.235,0.6873)
(0.24,0.6754)
(0.245,0.6634)
(0.25,0.6512)
(0.255,0.6389)
(0.26,0.6264)
(0.265,0.6139)
(0.27,0.6012)
(0.275,0.5884)
(0.28,0.5755)
(0.285,0.5625)
(0.29,0.5494)
(0.295,0.5362)
(0.3,0.523)
(0.305,0.5096)
(0.31,0.4962)
(0.315,0.4828)
(0.32,0.4693)
(0.325,0.4557)
(0.33,0.4421)
(0.335,0.4285)
(0.34,0.4148)
(0.345,0.4011)
(0.35,0.3874)
(0.355,0.3737)
(0.36,0.36)
(0.365,0.3462)
(0.37,0.3325)
(0.375,0.3188)
(0.38,0.3051)
(0.385,0.2915)
(0.39,0.2778)
(0.395,0.2642)
(0.4,0.2507)
(0.405,0.2372)
(0.41,0.2237)
(0.415,0.2103)
(0.42,0.197)
(0.425,0.1838)
(0.43,0.1706)
(0.435,0.1575)
(0.44,0.1445)
(0.445,0.1316)
(0.45,0.1188)
(0.455,0.1061)
(0.46,0.0935)
(0.465,0.0811)
(0.47,0.0688)
(0.475,0.0566)
(0.48,0.0445)
(0.485,0.0326)
(0.49,0.0208)
(0.495,0.0092)
(0.5,-0.0023)
(0.505,-0.0135)
(0.51,-0.0246)
(0.515,-0.0356)
(0.52,-0.0463)
(0.525,-0.0568)
(0.53,-0.0672)
(0.535,-0.0773)
(0.54,-0.0872)
(0.545,-0.0969)
(0.55,-0.1064)
(0.555,-0.1156)
(0.56,-0.1245)
(0.565,-0.1332)
(0.57,-0.1417)
(0.575,-0.1498)
(0.58,-0.1577)
(0.585,-0.1652)
(0.59,-0.1725)
(0.595,-0.1794)
(0.6,-0.186)
(0.605,-0.1922)
(0.61,-0.1981)
(0.615,-0.2035)
(0.62,-0.2086)
(0.625,-0.2133)
(0.63,-0.2175)
(0.635,-0.2213)
(0.64,-0.2246)
(0.645,-0.2275)
(0.65,-0.2299)
(0.655,-0.2317)
(0.66,-0.233)
(0.665,-0.2339)
(0.67,-0.2341)
(0.675,-0.2338)
(0.68,-0.233)
(0.685,-0.2316)
(0.69,-0.2297)
(0.695,-0.2272)
(0.7,-0.2242)
(0.705,-0.2207)
(0.71,-0.2166)
(0.715,-0.2121)
(0.72,-0.2071)
(0.725,-0.2017)
(0.73,-0.1958)
(0.735,-0.1896)
(0.74,-0.183)
(0.745,-0.176)
(0.75,-0.1688)
(0.755,-0.1612)
(0.76,-0.1535)
(0.765,-0.1454)
(0.77,-0.1372)
(0.775,-0.1289)
(0.78,-0.1203)
(0.785,-0.1117)
(0.79,-0.103)
(0.795,-0.0942)
(0.8,-0.0853)
(0.805,-0.0764)
(0.81,-0.0675)
(0.815,-0.0586)
(0.82,-0.0497)
(0.825,-0.0409)
(0.83,-0.0322)
(0.835,-0.0235)
(0.84,-0.0149)
(0.845,-0.0064)
(0.85,0.002)
(0.855,0.0102)
(0.86,0.0183)
(0.865,0.0263)
(0.87,0.0341)
(0.875,0.0417)
(0.88,0.0491)
(0.885,0.0563)
(0.89,0.0634)
(0.895,0.0702)
(0.9,0.0768)
(0.905,0.0832)
(0.91,0.0894)
(0.915,0.0953)
(0.92,0.101)
(0.925,0.1065)
(0.93,0.1117)
(0.935,0.1167)
(0.94,0.1215)
(0.945,0.126)
(0.95,0.1302)
(0.955,0.1342)
(0.96,0.1379)
(0.965,0.1414)
(0.97,0.1446)
(0.975,0.1476)
(0.98,0.1504)
(0.985,0.1528)
(0.99,0.1551)
(0.995,0.1571)
(1.0,0.1588)
(1.005,0.1604)
(1.01,0.1616)
(1.015,0.1627)
(1.02,0.1635)
(1.025,0.1641)
(1.03,0.1645)
(1.035,0.1647)
(1.04,0.1647)
(1.045,0.1644)
(1.05,0.164)
(1.055,0.1634)
(1.06,0.1626)
(1.065,0.1616)
(1.07,0.1605)
(1.075,0.1592)
(1.08,0.1577)
(1.085,0.1561)
(1.09,0.1544)
(1.095,0.1526)
(1.1,0.1506)
(1.105,0.1485)
(1.11,0.1463)
(1.115,0.144)
(1.12,0.1416)
(1.125,0.1392)
(1.13,0.1366)
(1.135,0.1341)
(1.14,0.1314)
(1.145,0.1288)
(1.15,0.1261)
(1.155,0.1234)
(1.16,0.1206)
(1.165,0.1179)
(1.17,0.1152)
(1.175,0.1125)
(1.18,0.1099)
(1.185,0.1073)
(1.19,0.1047)
(1.195,0.1022)
(1.2,0.0998)
(1.205,0.0974)
(1.21,0.0952)
(1.215,0.093)
(1.22,0.091)
(1.225,0.0891)
(1.23,0.0873)
(1.235,0.0856)
(1.24,0.0841)
(1.245,0.0827)
(1.25,0.0815)
(1.255,0.0805)
(1.26,0.0796)
(1.265,0.079)
(1.27,0.0785)
(1.275,0.0782)
(1.28,0.078)
(1.285,0.0781)
(1.29,0.0784)
(1.295,0.0789)
(1.3,0.0796)
(1.305,0.0805)
(1.31,0.0816)
(1.315,0.0829)
(1.32,0.0844)
(1.325,0.0861)
(1.33,0.088)
(1.335,0.0902)
(1.34,0.0925)
(1.345,0.0949)
(1.35,0.0976)
(1.355,0.1005)
(1.36,0.1035)
(1.365,0.1067)
(1.37,0.11)
(1.375,0.1135)
(1.38,0.1171)
(1.385,0.1209)
(1.39,0.1248)
(1.395,0.1288)
(1.4,0.1329)
(1.405,0.1372)
(1.41,0.1415)
(1.415,0.1459)
(1.42,0.1504)
(1.425,0.1549)
(1.43,0.1596)
(1.435,0.1642)
(1.44,0.1689)
(1.445,0.1737)
(1.45,0.1785)
(1.455,0.1833)
(1.46,0.1881)
(1.465,0.1929)
(1.47,0.1977)
(1.475,0.2025)
(1.48,0.2073)
(1.485,0.212)
(1.49,0.2168)
(1.495,0.2214)
(1.5,0.2261)
(1.505,0.2307)
(1.51,0.2352)
(1.515,0.2396)
(1.52,0.244)
(1.525,0.2483)
(1.53,0.2526)
(1.535,0.2567)
(1.54,0.2608)
(1.545,0.2647)
(1.55,0.2686)
(1.555,0.2723)
(1.56,0.2759)
(1.565,0.2794)
(1.57,0.2828)
(1.575,0.2861)
(1.58,0.2893)
(1.585,0.2923)
(1.59,0.2952)
(1.595,0.2979)
(1.6,0.3005)
(1.605,0.303)
(1.61,0.3053)
(1.615,0.3075)
(1.62,0.3096)
(1.625,0.3115)
(1.63,0.3132)
(1.635,0.3148)
(1.64,0.3163)
(1.645,0.3176)
(1.65,0.3188)
(1.655,0.3198)
(1.66,0.3207)
(1.665,0.3214)
(1.67,0.322)
(1.675,0.3225)
(1.68,0.3228)
(1.685,0.323)
(1.69,0.323)
(1.695,0.323)
(1.7,0.3228)
(1.705,0.3225)
(1.71,0.322)
(1.715,0.3215)
(1.72,0.3208)
(1.725,0.3201)
(1.73,0.3192)
(1.735,0.3182)
(1.74,0.3172)
(1.745,0.3161)
(1.75,0.3149)
(1.755,0.3136)
(1.76,0.3123)
(1.765,0.3109)
(1.77,0.3095)
(1.775,0.308)
(1.78,0.3065)
(1.785,0.305)
(1.79,0.3034)
(1.795,0.3018)
(1.8,0.3003)
(1.805,0.2987)
(1.81,0.2971)
(1.815,0.2956)
(1.82,0.294)
(1.825,0.2925)
(1.83,0.291)
(1.835,0.2896)
(1.84,0.2882)
(1.845,0.2869)
(1.85,0.2856)
(1.855,0.2843)
(1.86,0.2831)
(1.865,0.282)
(1.87,0.2809)
(1.875,0.2799)
(1.88,0.279)
(1.885,0.2781)
(1.89,0.2773)
(1.895,0.2766)
(1.9,0.2759)
(1.905,0.2753)
(1.91,0.2747)
(1.915,0.2742)
(1.92,0.2737)
(1.925,0.2733)
(1.93,0.2729)
(1.935,0.2726)
(1.94,0.2722)
(1.945,0.272)
(1.95,0.2717)
(1.955,0.2715)
(1.96,0.2713)
(1.965,0.2711)
(1.97,0.2709)
(1.975,0.2708)
(1.98,0.2706)
(1.985,0.2704)
(1.99,0.2702)
(1.995,0.27)
(2.0,0.2698)
(2.005,0.2696)
(2.01,0.2694)
(2.015,0.2692)
(2.02,0.2689)
(2.025,0.2686)
(2.03,0.2683)
(2.035,0.2679)
(2.04,0.2675)
(2.045,0.2671)
(2.05,0.2667)
(2.055,0.2662)
(2.06,0.2657)
(2.065,0.2651)
(2.07,0.2645)
(2.075,0.2639)
(2.08,0.2633)
(2.085,0.2626)
(2.09,0.2619)
(2.095,0.2611)
(2.1,0.2604)
(2.105,0.2596)
(2.11,0.2587)
(2.115,0.2579)
(2.12,0.257)
(2.125,0.2561)
(2.13,0.2552)
(2.135,0.2542)
(2.14,0.2533)
(2.145,0.2523)
(2.15,0.2513)
(2.155,0.2504)
(2.16,0.2494)
(2.165,0.2484)
(2.17,0.2474)
(2.175,0.2464)
(2.18,0.2454)
(2.185,0.2445)
(2.19,0.2435)
(2.195,0.2426)
(2.2,0.2417)
(2.205,0.2408)
(2.21,0.24)
(2.215,0.2392)
(2.22,0.2384)
(2.225,0.2376)
(2.23,0.2369)
(2.235,0.2363)
(2.24,0.2357)
(2.245,0.2351)
(2.25,0.2346)
(2.255,0.2341)
(2.26,0.2337)
(2.265,0.2333)
(2.27,0.233)
(2.275,0.2328)
(2.28,0.2326)
(2.285,0.2324)
(2.29,0.2323)
(2.295,0.2323)
(2.3,0.2323)
(2.305,0.2324)
(2.31,0.2325)
(2.315,0.2326)
(2.32,0.2328)
(2.325,0.2331)
(2.33,0.2334)
(2.335,0.2337)
(2.34,0.234)
(2.345,0.2344)
(2.35,0.2348)
(2.355,0.2352)
(2.36,0.2356)
(2.365,0.236)
(2.37,0.2365)
(2.375,0.2369)
(2.38,0.2373)
(2.385,0.2377)
(2.39,0.2381)
(2.395,0.2384)
(2.4,0.2388)
(2.405,0.2391)
(2.41,0.2393)
(2.415,0.2395)
(2.42,0.2397)
(2.425,0.2398)
(2.43,0.2399)
(2.435,0.2399)
(2.44,0.2398)
(2.445,0.2397)
(2.45,0.2395)
(2.455,0.2392)
(2.46,0.2389)
(2.465,0.2385)
(2.47,0.2379)
(2.475,0.2373)
(2.48,0.2367)
(2.485,0.2359)
(2.49,0.235)
(2.495,0.234)
(2.5,0.233)};
\addplot[name path=positive_down,color=blue!70] coordinates {(0.005,0.9772)
(0.01,0.9765)
(0.015,0.9753)
(0.02,0.9735)
(0.025,0.9713)
(0.03,0.9684)
(0.035,0.965)
(0.04,0.9609)
(0.045,0.9561)
(0.05,0.9506)
(0.055,0.9444)
(0.06,0.9374)
(0.065,0.9296)
(0.07,0.921)
(0.075,0.9117)
(0.08,0.9017)
(0.085,0.8909)
(0.09,0.8794)
(0.095,0.8672)
(0.1,0.8543)
(0.105,0.8408)
(0.11,0.8267)
(0.115,0.812)
(0.12,0.7968)
(0.125,0.7809)
(0.13,0.7645)
(0.135,0.7476)
(0.14,0.7302)
(0.145,0.7123)
(0.15,0.694)
(0.155,0.6752)
(0.16,0.656)
(0.165,0.6363)
(0.17,0.6163)
(0.175,0.5959)
(0.18,0.5751)
(0.185,0.5541)
(0.19,0.5327)
(0.195,0.511)
(0.2,0.4891)
(0.205,0.4669)
(0.21,0.4444)
(0.215,0.4218)
(0.22,0.399)
(0.225,0.376)
(0.23,0.3529)
(0.235,0.3296)
(0.24,0.3063)
(0.245,0.2829)
(0.25,0.2594)
(0.255,0.2358)
(0.26,0.2123)
(0.265,0.1887)
(0.27,0.1652)
(0.275,0.1417)
(0.28,0.1182)
(0.285,0.0949)
(0.29,0.0716)
(0.295,0.0485)
(0.3,0.0255)
(0.305,0.0027)
(0.31,-0.02)
(0.315,-0.0425)
(0.32,-0.0647)
(0.325,-0.0867)
(0.33,-0.1085)
(0.335,-0.13)
(0.34,-0.1512)
(0.345,-0.1722)
(0.35,-0.1928)
(0.355,-0.213)
(0.36,-0.233)
(0.365,-0.2526)
(0.37,-0.2718)
(0.375,-0.2906)
(0.38,-0.309)
(0.385,-0.327)
(0.39,-0.3446)
(0.395,-0.3618)
(0.4,-0.3785)
(0.405,-0.3948)
(0.41,-0.4106)
(0.415,-0.4259)
(0.42,-0.4408)
(0.425,-0.4551)
(0.43,-0.469)
(0.435,-0.4824)
(0.44,-0.4953)
(0.445,-0.5076)
(0.45,-0.5195)
(0.455,-0.5308)
(0.46,-0.5416)
(0.465,-0.5519)
(0.47,-0.5617)
(0.475,-0.5709)
(0.48,-0.5797)
(0.485,-0.5879)
(0.49,-0.5956)
(0.495,-0.6027)
(0.5,-0.6094)
(0.505,-0.6155)
(0.51,-0.6212)
(0.515,-0.6263)
(0.52,-0.631)
(0.525,-0.6351)
(0.53,-0.6388)
(0.535,-0.642)
(0.54,-0.6448)
(0.545,-0.6471)
(0.55,-0.649)
(0.555,-0.6505)
(0.56,-0.6515)
(0.565,-0.6522)
(0.57,-0.6524)
(0.575,-0.6523)
(0.58,-0.6519)
(0.585,-0.6511)
(0.59,-0.6501)
(0.595,-0.6487)
(0.6,-0.6471)
(0.605,-0.6453)
(0.61,-0.6432)
(0.615,-0.6409)
(0.62,-0.6385)
(0.625,-0.6359)
(0.63,-0.6333)
(0.635,-0.6305)
(0.64,-0.6277)
(0.645,-0.6249)
(0.65,-0.6221)
(0.655,-0.6193)
(0.66,-0.6165)
(0.665,-0.6139)
(0.67,-0.6113)
(0.675,-0.6089)
(0.68,-0.6066)
(0.685,-0.6045)
(0.69,-0.6025)
(0.695,-0.6008)
(0.7,-0.5992)
(0.705,-0.5978)
(0.71,-0.5966)
(0.715,-0.5956)
(0.72,-0.5947)
(0.725,-0.5941)
(0.73,-0.5936)
(0.735,-0.5932)
(0.74,-0.593)
(0.745,-0.5929)
(0.75,-0.5929)
(0.755,-0.593)
(0.76,-0.5932)
(0.765,-0.5934)
(0.77,-0.5938)
(0.775,-0.5941)
(0.78,-0.5945)
(0.785,-0.5949)
(0.79,-0.5954)
(0.795,-0.5958)
(0.8,-0.5962)
(0.805,-0.5967)
(0.81,-0.597)
(0.815,-0.5974)
(0.82,-0.5977)
(0.825,-0.598)
(0.83,-0.5983)
(0.835,-0.5985)
(0.84,-0.5986)
(0.845,-0.5987)
(0.85,-0.5987)
(0.855,-0.5987)
(0.86,-0.5986)
(0.865,-0.5984)
(0.87,-0.5981)
(0.875,-0.5978)
(0.88,-0.5974)
(0.885,-0.597)
(0.89,-0.5964)
(0.895,-0.5958)
(0.9,-0.5951)
(0.905,-0.5944)
(0.91,-0.5935)
(0.915,-0.5926)
(0.92,-0.5916)
(0.925,-0.5905)
(0.93,-0.5894)
(0.935,-0.5882)
(0.94,-0.5868)
(0.945,-0.5855)
(0.95,-0.584)
(0.955,-0.5825)
(0.96,-0.5809)
(0.965,-0.5792)
(0.97,-0.5775)
(0.975,-0.5756)
(0.98,-0.5738)
(0.985,-0.5718)
(0.99,-0.5698)
(0.995,-0.5677)
(1.0,-0.5655)
(1.005,-0.5633)
(1.01,-0.561)
(1.015,-0.5586)
(1.02,-0.5562)
(1.025,-0.5537)
(1.03,-0.5512)
(1.035,-0.5486)
(1.04,-0.546)
(1.045,-0.5433)
(1.05,-0.5405)
(1.055,-0.5377)
(1.06,-0.5348)
(1.065,-0.5319)
(1.07,-0.5289)
(1.075,-0.5259)
(1.08,-0.5229)
(1.085,-0.5198)
(1.09,-0.5166)
(1.095,-0.5134)
(1.1,-0.5102)
(1.105,-0.507)
(1.11,-0.5037)
(1.115,-0.5004)
(1.12,-0.497)
(1.125,-0.4937)
(1.13,-0.4903)
(1.135,-0.4869)
(1.14,-0.4835)
(1.145,-0.48)
(1.15,-0.4766)
(1.155,-0.4731)
(1.16,-0.4696)
(1.165,-0.4662)
(1.17,-0.4627)
(1.175,-0.4592)
(1.18,-0.4558)
(1.185,-0.4523)
(1.19,-0.4489)
(1.195,-0.4455)
(1.2,-0.4421)
(1.205,-0.4387)
(1.21,-0.4354)
(1.215,-0.4321)
(1.22,-0.4289)
(1.225,-0.4256)
(1.23,-0.4225)
(1.235,-0.4193)
(1.24,-0.4163)
(1.245,-0.4133)
(1.25,-0.4103)
(1.255,-0.4074)
(1.26,-0.4046)
(1.265,-0.4019)
(1.27,-0.3992)
(1.275,-0.3966)
(1.28,-0.394)
(1.285,-0.3916)
(1.29,-0.3892)
(1.295,-0.3868)
(1.3,-0.3846)
(1.305,-0.3824)
(1.31,-0.3803)
(1.315,-0.3783)
(1.32,-0.3763)
(1.325,-0.3744)
(1.33,-0.3726)
(1.335,-0.3708)
(1.34,-0.3691)
(1.345,-0.3674)
(1.35,-0.3658)
(1.355,-0.3642)
(1.36,-0.3626)
(1.365,-0.3611)
(1.37,-0.3596)
(1.375,-0.3581)
(1.38,-0.3566)
(1.385,-0.3552)
(1.39,-0.3537)
(1.395,-0.3523)
(1.4,-0.3508)
(1.405,-0.3493)
(1.41,-0.3478)
(1.415,-0.3463)
(1.42,-0.3448)
(1.425,-0.3432)
(1.43,-0.3416)
(1.435,-0.34)
(1.44,-0.3383)
(1.445,-0.3366)
(1.45,-0.3348)
(1.455,-0.3329)
(1.46,-0.331)
(1.465,-0.3291)
(1.47,-0.3271)
(1.475,-0.325)
(1.48,-0.3229)
(1.485,-0.3206)
(1.49,-0.3184)
(1.495,-0.316)
(1.5,-0.3136)
(1.505,-0.3111)
(1.51,-0.3086)
(1.515,-0.3059)
(1.52,-0.3032)
(1.525,-0.3004)
(1.53,-0.2976)
(1.535,-0.2947)
(1.54,-0.2917)
(1.545,-0.2886)
(1.55,-0.2855)
(1.555,-0.2823)
(1.56,-0.279)
(1.565,-0.2757)
(1.57,-0.2723)
(1.575,-0.2689)
(1.58,-0.2653)
(1.585,-0.2618)
(1.59,-0.2582)
(1.595,-0.2545)
(1.6,-0.2508)
(1.605,-0.247)
(1.61,-0.2432)
(1.615,-0.2393)
(1.62,-0.2354)
(1.625,-0.2315)
(1.63,-0.2276)
(1.635,-0.2236)
(1.64,-0.2196)
(1.645,-0.2155)
(1.65,-0.2115)
(1.655,-0.2074)
(1.66,-0.2034)
(1.665,-0.1993)
(1.67,-0.1953)
(1.675,-0.1912)
(1.68,-0.1872)
(1.685,-0.1831)
(1.69,-0.1791)
(1.695,-0.1751)
(1.7,-0.1712)
(1.705,-0.1673)
(1.71,-0.1634)
(1.715,-0.1596)
(1.72,-0.1558)
(1.725,-0.1521)
(1.73,-0.1484)
(1.735,-0.1448)
(1.74,-0.1413)
(1.745,-0.1378)
(1.75,-0.1345)
(1.755,-0.1312)
(1.76,-0.128)
(1.765,-0.1249)
(1.77,-0.122)
(1.775,-0.1191)
(1.78,-0.1164)
(1.785,-0.1138)
(1.79,-0.1113)
(1.795,-0.109)
(1.8,-0.1068)
(1.805,-0.1047)
(1.81,-0.1028)
(1.815,-0.101)
(1.82,-0.0994)
(1.825,-0.098)
(1.83,-0.0967)
(1.835,-0.0955)
(1.84,-0.0946)
(1.845,-0.0938)
(1.85,-0.0932)
(1.855,-0.0927)
(1.86,-0.0924)
(1.865,-0.0923)
(1.87,-0.0923)
(1.875,-0.0925)
(1.88,-0.0928)
(1.885,-0.0933)
(1.89,-0.094)
(1.895,-0.0948)
(1.9,-0.0957)
(1.905,-0.0967)
(1.91,-0.0979)
(1.915,-0.0992)
(1.92,-0.1006)
(1.925,-0.1021)
(1.93,-0.1037)
(1.935,-0.1053)
(1.94,-0.1071)
(1.945,-0.1089)
(1.95,-0.1108)
(1.955,-0.1127)
(1.96,-0.1147)
(1.965,-0.1167)
(1.97,-0.1188)
(1.975,-0.1208)
(1.98,-0.1229)
(1.985,-0.1249)
(1.99,-0.127)
(1.995,-0.1291)
(2.0,-0.1311)
(2.005,-0.1331)
(2.01,-0.1351)
(2.015,-0.137)
(2.02,-0.1389)
(2.025,-0.1408)
(2.03,-0.1426)
(2.035,-0.1444)
(2.04,-0.146)
(2.045,-0.1477)
(2.05,-0.1492)
(2.055,-0.1507)
(2.06,-0.1521)
(2.065,-0.1535)
(2.07,-0.1547)
(2.075,-0.1559)
(2.08,-0.157)
(2.085,-0.158)
(2.09,-0.159)
(2.095,-0.1599)
(2.1,-0.1606)
(2.105,-0.1613)
(2.11,-0.162)
(2.115,-0.1625)
(2.12,-0.163)
(2.125,-0.1634)
(2.13,-0.1637)
(2.135,-0.1639)
(2.14,-0.1641)
(2.145,-0.1642)
(2.15,-0.1643)
(2.155,-0.1643)
(2.16,-0.1642)
(2.165,-0.1641)
(2.17,-0.164)
(2.175,-0.1638)
(2.18,-0.1636)
(2.185,-0.1633)
(2.19,-0.163)
(2.195,-0.1627)
(2.2,-0.1624)
(2.205,-0.1621)
(2.21,-0.1617)
(2.215,-0.1614)
(2.22,-0.1611)
(2.225,-0.1608)
(2.23,-0.1604)
(2.235,-0.1602)
(2.24,-0.1599)
(2.245,-0.1597)
(2.25,-0.1595)
(2.255,-0.1593)
(2.26,-0.1592)
(2.265,-0.1591)
(2.27,-0.1591)
(2.275,-0.1591)
(2.28,-0.1592)
(2.285,-0.1593)
(2.29,-0.1595)
(2.295,-0.1598)
(2.3,-0.1601)
(2.305,-0.1605)
(2.31,-0.161)
(2.315,-0.1615)
(2.32,-0.1621)
(2.325,-0.1628)
(2.33,-0.1635)
(2.335,-0.1643)
(2.34,-0.1651)
(2.345,-0.166)
(2.35,-0.167)
(2.355,-0.168)
(2.36,-0.1691)
(2.365,-0.1702)
(2.37,-0.1714)
(2.375,-0.1726)
(2.38,-0.1739)
(2.385,-0.1752)
(2.39,-0.1765)
(2.395,-0.1779)
(2.4,-0.1793)
(2.405,-0.1808)
(2.41,-0.1822)
(2.415,-0.1837)
(2.42,-0.1852)
(2.425,-0.1867)
(2.43,-0.1882)
(2.435,-0.1897)
(2.44,-0.1912)
(2.445,-0.1927)
(2.45,-0.1942)
(2.455,-0.1957)
(2.46,-0.1972)
(2.465,-0.1987)
(2.47,-0.2001)
(2.475,-0.2015)
(2.48,-0.2029)
(2.485,-0.2043)
(2.49,-0.2057)
(2.495,-0.207)
(2.5,-0.2083)};
\addplot[blue!50,fill opacity=0.5] fill between[of=positive_top and positive_down];

\legend{Popular pages,,,, Unpopular pages}
\end{axis}
\end{tikzpicture}}

\caption{The real part of class dependent mean characteristic functions with standard deviations around the mean for the log transformed degree on the Wikipedia Crocodiles dataset.}\label{fig:selling}
\end{figure}

%% file: sections/related_work.tex
\section{Related work}\label{sec:feather_related_work}

Characteristic functions have previously been used in relation to heat diffusion wavelets~\cite{graphwave}, which defined the functions for uniform ties strengths and restricted features types. 

\textit{Node embedding} techniques map nodes of a graph into Euclidean spaces where the similarity of vertices is approximately preserved -- each node has a vector representation. Various forms of embeddings have been studied recently,   \textit{Neighbourhood preserving} node embeddings  are learned by explicitly \citep{netmf, hope, grarep} or implicitly decomposing \cite{deepwalk, walklets, line} a proximity matrix of the graph. \textit{Attributed node embedding} techniques  \cite{asne, sine, tadw, tene, bane} augment the neighbourhood information with generic node attributes (e.g. the user's age in a social network) and nodes sharing metadata are closer in the learned embedding space. \textit{Structural embeddings} \cite{role2vec,rolx,node2vec} create vector representations of nodes which retain the similarity of structural roles and equivalences of nodes. The non-supervised \textit{FEATHER} algorithm which we propose can be seen as a node embedding technique. We create a mapping of nodes to the Euclidean space, simply by evaluating the characteristic function for metadata based generic, neighbourhood and structural node attributes. With the appropriate tie strength definition we are able to hybridize all three types of information with our embedding.

\textit{Whole graph embedding} and \textit{statistical graph fingerprinting} techniques map graphs into Euclidean spaces where graphs with similar structures and subpatterns are located in close proximity -- each graph obtains a vector representation. Whole graph embedding procedures \cite{graph2vec, gl2vec} achieve this by decomposing graph -- structural feature matrices to learn an embedding. Statistical graph fingerprinting techniques \cite{fgsd, sf, netlsd, geoscattering} extract information from the graph Laplacian eigenvalues, or using the graph scattering transform without learning. Node level \textit{FEATHER} representations can be pooled by permutation invariant functions to output condensed graph fingerprints which is in principle similar to statistical graph fingerprinting. These statistical fingerprints are related to \textit{graph kernels} as the pooled characteristic functions can serve as inputs for appropriate kernel functions. This way the similarity of graphs is not compared based on the presence of sparsely appearing common random walks \cite{rwkernel}, cyclic patterns \cite{cyclickernel} or subtree patterns \cite{wlkernel}, but via the kernel defined on pairs of dense pooled graph characteristic function representations.

There is also a parallel between \textit{FEATHER} and the forward pass of \textit{graph neural network layers} \cite{gcn, graphsage}. During the \textit{FEATHER} function evaluation using the tie strength weights and vertex features we create multi-scale descriptors of the feature distributions which are parameterized by the evaluation points. This can be seen as the forward pass of a multi-scale graph neural network \cite{mixhop, appnp} which is able to describe vertex features at multiple scales. Using this we essentially define end-to-end differentiable parametric statistical models where the modulation of evaluation points (the relaxation of fixed evaluation points) can help with the downstream learning task at hand. Compared to traditional graph neural networks \cite{gcn, graphsage, sgconv, appnp, mixhop,clustergcn}, which only calculate the first moments of the node feature distributions, \textit{FEATHER} models give summaries of node feature distributions with trainable characteristic function evaluation points.

%% file: sections/characteristic_function.tex
\input{figures/groupped_char.tex}
\section{Characteristic functions on graphs}\label{sec:feather_theory}
In this section we introduce the idea of characteristic functions defined on attributed graphs. Specifically, we discuss the idea of describing node feature distributions in a neighbourhood with characteristic functions. We propose a specific instance of these functions, the \textit{r-scale random walk weighted characteristic function} and we describe an algorithm to calculate this function for all nodes in linear time. We prove the robustness of these functions and how they represent isomorphic graphs when node level functions are pooled. Finally, we discuss how characteristic functions can serve as building blocks for parametric statistical models.

\subsection{Node feature distribution characterization}\label{subsec:runtime}

We assume that we have an attributed and undirected graph $G=(V,E)$. Nodes of $G$ have a feature described by the random variable $X$, specifically defined as the feature vector $\textbf{x}\in \mathbb{R}^{|V|}$, where $\textbf{x}_v$ is the feature value for node $v\in V$. We are interested in describing the distribution of this feature in the neighbourhood of $u \in V$. 
The characteristic function of $X$ for source node $u$ at characteristic function evaluation point $\theta\in \mathbb{R}$ is the function defined by Equation \eqref{eq:basic_definition} where $i$ denotes the imaginary unit.
\begin{align}
\operatorname{E}\left[e^{i\theta X}|G,u\right] &= \sum \limits_{w \in V}P(w|u) \cdot e^{i\theta\textbf{x}_w}\label{eq:basic_definition}
\end{align}
In Equation \eqref{eq:basic_definition} the \textit{affiliation probability} $P(w|u)$ describes the strength of the relationship between the source node $u$ and the target node $w \in V$. We would like to emphasize that the source node $u$ and the target nodes do not have to be connected directly and that $\sum_{w \in V}P(w|u)=1$ holds $\forall u \in V$. We use Euler's identity to obtain the real and imaginary part of the function described by Equation \eqref{eq:basic_definition} which are respectively defined by Equations \eqref{eq:cos_with_prob_basic} and \eqref{eq:sin_with_prob_basic}.
\begin{align}
\text{Re}\left(\operatorname{E}\left[e^{i\theta X}|G, u\right]\right)&=\sum \limits_{w \in V}P(w|u)\cos(\theta\textbf{x}_w)\label{eq:cos_with_prob_basic}\\
\text{Im}\left(\operatorname{E}\left[e^{i\theta X}|G, u\right]\right)&=\sum \limits_{w \in V}P(w|u)\sin(\theta\textbf{x}_w)\label{eq:sin_with_prob_basic}
\end{align}
The real and imaginary parts of the characteristic function are respectively weighted sums of cosine and sine waves where the weight of an individual wave is $P(w|u)$, the evaluation point $\theta$ is equivalent to time, and $\textbf{x}_w$ describes the angular frequency.

\subsubsection{The r-scale random walk weighted characteristic function}
So far we have not specified how the affiliation probability $P(w|u)$ between the source $u$ and target $w$ is parametrized. Now we will introduce a parametrization which uses random walk transition probabilities. The sequence of nodes in a random walk on $G$ is denoted by $\{ v_j,v_{j+1},\dots, v_{j+r}\}$.

Let us assume that the neighbourhood of $u$ at scale $r$ consists of nodes that can be reached by a random walk in $r$ steps from source node $u$. We are interested in describing the distribution of the feature in the neighbourhood of $u \in V$ at scale $r$ with the real and imaginary parts of the characteristic function -- these are respectively defined by Equations \eqref{eq:cos_with_prob_r_scale} and \eqref{eq:sin_with_prob_r_scale}. 

\begin{align}
\text{Re}\left(\operatorname{E}\left[e^{i\theta X}|G,u, r\right]\right)&=\sum \limits_{w \in V}P(v_{j+r}=w|v_j=u)\cos(\theta\textbf{x}_w)\label{eq:cos_with_prob_r_scale}\\
\text{Im}\left(\operatorname{E}\left[e^{i\theta X}|G,u,r\right]\right)&=\sum \limits_{w \in V}P(v_{j+r}=w|v_j=u)\sin(\theta\textbf{x}_w)\label{eq:sin_with_prob_r_scale}
\end{align}

In Equations \eqref{eq:cos_with_prob_r_scale} and \eqref{eq:sin_with_prob_r_scale},  $P(v_{j+r}=w|v_j=u)=P(w|u)$ is the probability of a random walk starting from source node $u$, hitting the target node $w$ in the $r^{th}$ step. The adjacency matrix of $G$ is denoted by $\textbf{A}$ and the weighted diagonal degree matrix is $\textbf{D}$. The normalized adjacency matrix is defined as $\widehat{\textbf{A}}=\textbf{D}^{-1} \textbf{A}$. We can exploit the fact that, for a source-target node pair $(u,w)$ and a scale $r$, we can express $P(v_{j+r}=w|v_j=u)$ with the $r^{th}$ power of the normalized adjacency matrix. Using $\widehat{\textbf{A}}^r_{u,w}=P(v_{j+r}=w|v_j=u)$, we get Equations \eqref{eq:cos_mat_identity} and \eqref{eq:sin_mat_identity}.
\begin{align}
\text{Re}\left(\operatorname{E}\left[e^{i\theta X}|G,u, r\right]\right)&=\sum \limits_{w \in V}\widehat{\textbf{A}}^r_{u,w}\cos(\theta\textbf{x}_w)\label{eq:cos_mat_identity}\\
\text{Im}\left(\operatorname{E}\left[e^{i\theta X}|G,u,r\right]\right)&=\sum \limits_{w \in V}\widehat{\textbf{A}}^r_{u,w}\sin(\theta \textbf{x}_w)\label{eq:sin_mat_identity}
\end{align}

Figure \ref{fig:characteristic_function_scales} shows the real and imaginary part of the \textit{r-scale random walk weighted characteristic function} of the log transformed degree for a low and high degree node in the Twitch England network \cite{musae}. A few important properties of the function are visible; (i) the real part is an even function while the imaginary part is odd, (ii) the range of both parts is in the [-1,1] interval, (iii) nodes with different structural roles have different characteristic functions.

\subsubsection{Efficient calculation of the \textit{r}-scale random walk weighted characteristic function}
Up to this point we have only discussed the characteristic function at scale $r$ for a single $u\in V$. However, we might want to characterize every node with respect to a feature in the graph in an efficient way.
Moreover, we do not want to evaluate each node characteristic function on the whole domain. Because of this we will sample $d$ points from the domain and evaluate the function at these points which are described by the \textit{evaluation point vector} $\Theta\in \mathbb{R}^d$.
We define the \textit{r}-scale random walk weighted characteristic function of the whole graph as the complex matrix valued function denoted as $\mathcal{CF}(G,X,\Theta,r)\rightarrow \mathbb{C}^{|V|\times d}$.   
The real and imaginary parts of this complex matrix valued function are described by the matrix valued functions in Equations \eqref{eq:matrix_real} and \eqref{eq:matrix_imag} respectively.
\begin{align}
\text{Re}(\mathcal{CF}(G,X,\Theta,r))&=\widehat{\textbf{A}}^r\cdot \cos(\textbf{x} \otimes \Theta)\label{eq:matrix_real}\\
\text{Im}(\mathcal{CF}(G,X,\Theta,r))&=\widehat{\textbf{A}}^r\cdot \sin(\textbf{x} \otimes \Theta)\label{eq:matrix_imag}
\end{align}
These matrices describe the feature distributions around nodes if two rows are similar it, implies that the corresponding nodes have similar distributions of the feature around them at scale \textit{r}. This representation can be seen as a \textit{node embedding}, which characterizes the nodes in terms of the local feature distribution. Calculating the \textit{r}-scale random walk weighted characteristic function for the whole graph has a time complexity of $\mathcal{O}(|E|\cdot d \cdot r)$ and memory complexity of $\mathcal{O}(|V|\cdot d)$.

\subsubsection{Characterizing multiple features for all nodes}

Up to this point, we have assumed that the nodes have a single feature, described by the feature vector $\textbf{x}\in \mathbb{R}^{|V|}$. Now we will consider the more general case when we have a set of $k$ node feature vectors. In a social network these vectors might describe the age, income, and other generic real valued properties of the users. This set of vertex features is defined by $\mathcal{X}=\{\textbf{x}^1,  \dots, \textbf{x}^k\}$.

We now consider the design of an efficient sparsity aware algorithm which can calculate the $r$ scale random walk weighted characteristic function for each node and feature. We named this procedure \textit{FEATHER}, and it is summarized by Algorithm \ref{characteristic_function_algo}. It evaluates the characteristic functions for a graph for each feature $\textbf{x} \in \mathcal{X}$ at all scales up to $r$. The connectivity of the graph is described by the normalized adjacency matrix $\widehat{\textbf{A}}$. For each feature vector $\textbf{x}^i,\,\,i\in{1,\dots,k}$ at scale $r$ we have a corresponding characteristic function evaluation vector $\Theta^{i,r}\in\mathbb{R}^d$. For simplicity we assume that we evaluate the characteristic functions at the same number of points.

Let us look at the mechanics of Algorithm \ref{characteristic_function_algo}. First, we initialize the real and imaginary parts of the embeddings denoted by $\textbf{Z}_{Re}$ and $\textbf{Z}_{Im}$ respectively (lines 1 and 2). We iterate over the $k$ different node features (line 3) and the scales up to $r$ (line 4). When we consider the first scale (line 6) we calculate the outer product of the feature being considered and the corresponding evaluation point vector -- this results in $\textbf{H}$ (line 7). We elementwise take the sine and cosine of this matrix (lines 8 and 9). For each scale we calculate the real and imaginary parts of the graph characteristic function evaluations ($\textbf{H}_{Re}$ and $\textbf{H}_{Im}$) -- we use the normalized adjacency matrix to define the probability weights (lines 10 and 11). We append these matrices to the real and imaginary part of the embeddings (lines 13 and 14). When the characteristic function of each feature is evaluated at every scale we concatenate the real and imaginary part of the embeddings (line 17) and we return this embedding (line 18).

\input{./algorithms/r_scale_characteristic_function_calculation.tex}

Calculating the outer product (line 7) $\textbf{H}$ takes $O(|V|\cdot d)$ memory and time. The probability weighting (lines 10 and 11) is an operation which requires  $O(|V|\cdot d)$ memory and $O(|E| \cdot d)$ time. We do this for each feature at each scale with a separate evaluation point vector. This means that altogether calculating the $r$ scale graph characteristic function for each feature has a time complexity of $\mathcal{O}((|E|+|V|)\cdot d \cdot r^2 \cdot k )$ and the memory complexity of storing the embedding is $\mathcal{O}(|V|\cdot d \cdot r \cdot k)$.
\subsection{Theoretical properties}
We focus on two theoretical aspects of the $r$-scale random weighted characteristic function which have practical implications: robustness and how it represents isomorphic graphs. 
\begin{remark}
Let us consider a graph $G$, the feature $X$ and its corrupted variant $X'$ represented by the vectors $\textbf{x}$ and $\textbf{x}'$. The corrupted feature vector only differs from \textbf{x} at a single node $w\in V$ where $\textbf{x}'_w = \textbf{x}_w\pm\varepsilon$ for any $\varepsilon\in\mathbb{R}$. The absolute changes in the real and imaginary parts of the $r$-scale random walk weighted characteristic function for any $u\in V$ and $\theta\in\mathbb{R}$ satisfy that: 
\end{remark}
\begin{align*}
\underbrace{\left|\text{Re}\left(\operatorname{E}\left[e^{i\theta X}|G,u, r\right]\right)-\text{Re}\left(\operatorname{E}\left[e^{i\theta X'}|G,u, r\right]\right)\right|}_{\Delta\text{Re}}&\leq 2\cdot \widehat{\textbf{A}}^r_{u,w}\\
\underbrace{\left|\text{Im}\left(\operatorname{E}\left[e^{i\theta X}|G,u, r\right]\right)-\text{Im}\left(\operatorname{E}\left[e^{i\theta X'}|G,u, r\right]\right)\right|}_{\Delta\text{Im}}&\leq 2\cdot\widehat{\textbf{A}}^r_{u,w}.
\end{align*}
\begin{proof}
We know that the absolute difference in the real and imaginary part of the characteristic function is bounded by the maxima of such differences:
$$|\Delta \text{Re}|  \leq \max |\Delta \text{Re}|\quad\text{ and }\quad|\Delta \text{Im}| \leq \max |\Delta \text{Im}|.$$
We will prove the bound for the real part, the proof for the imaginary one follows similarly. Let us substitute the difference of the characteristic functions in the right hand side of the bound:
$$|\Delta \text{Re}|  \leq \max \left |\sum \limits_{v \in V}\widehat{\textbf{A}}^r_{u,v}\cos(\theta\textbf{x}_v)-\sum \limits_{v \in V}\widehat{\textbf{A}}^r_{u,v}\cos(\theta\textbf{x}'_v) \right |.$$
We exploit that $x_v=x_v',\,\,\forall v \in V\setminus \{w\}$ so we can rewrite the difference of sums because $ \cos(\theta\textbf{x}_v)-\cos(\theta\textbf{x}'_v)=0,\,\,\forall v \in V\setminus \{w\}$.

{\scriptsize
\begin{align*}
 |\Delta \text{Re}|  &\leq\max \left |\sum\limits_{v \in V \setminus \{ w\}}\left [ \widehat{\textbf{A}}^r_{u,v}\underbrace{\left(\cos(\theta\textbf{x}_v)-\cos(\theta\textbf{x}'_v)\right)}_{0}\right ]+\widehat{\textbf{A}}^r_{u,w}\left(\cos(\theta\textbf{x}_w)-\cos(\theta\textbf{x}'_w)\right)\right|
 \end{align*}
 }

The maximal absolute difference between two cosine functions is 2 so our proof is complete which means that the effect of corrupted features on the r-scale random walk weighted characteristic function values is bounded by the tie strength regardless the extent of data corruption. 
\begin{align*}
|\Delta \text{Re}|&\leq \widehat{\textbf{A}}^r_{u,w}\cdot \underbrace{\max \left|\cos(\theta\textbf{x}_w)-\cos(\theta\textbf{x}'_w)\right|}_{2}
\end{align*}
\end{proof}

\begin{definition}\label{def:pooling}
The real and imaginary part of the \textbf{mean pooled} $r$-scale random walk weighted characteristic function are defined as 
 $\sum\limits_{u\in V}  \sum \limits_{w \in V}\widehat{\textbf{A}}^r_{u,w}\cos(\theta\textbf{x}_w)/|V|$ and  $\sum\limits_{u\in V}  \sum \limits_{w \in V}\widehat{\textbf{A}}^r_{u,w}\sin(\theta\textbf{x}_w)/|V|$.
\end{definition}
The functions described by Definition \ref{def:pooling} allow for the characterization and comparison of whole graphs based on structural properties. Moreover, these descriptors can serve as features for graph level machine learning algorithms.
\begin{remark}
Given two isomorphic graphs $G,G'$ and the respective degree vectors $\textbf{x},\textbf{x}'$ the mean pooled $r$-scale random walk weighted degree characteristic functions are the same.
\end{remark}
\begin{proof}
Let us denote the normalized adjacency matrices of $G$ and $G$' as $\widehat{\textbf{A}}$ and $\widehat{\textbf{A}}'$. Because $G$ and $G'$ are isomorphic there is a $\textbf{P}$ permutation matrix for which it holds that $\widehat{\textbf{A}} = \textbf{P} \widehat{\textbf{A}}' \textbf{P}^{-1}.$ Using the same permutation matrix we get that $\textbf{x}=\textbf{P}\textbf{x}'$. Using Definition \ref{def:pooling} and the previous two equations it follows that the real and imaginary parts of pooled characteristic functions satisfy that
\begin{align*}
 \sum\limits_{u\in V}  \sum \limits_{w \in V}\widehat{\textbf{A}}^r_{u,w}\cos(\theta\textbf{x}_w)/|V|&= \sum\limits_{u\in V}  \sum \limits_{w \in V}(\textbf{P} \widehat{\textbf{A}} \textbf{P}^{-1})^r_{u,w}\cos(\theta\cdot(\textbf{P} \textbf{x})_w)/|V|\\
 \sum\limits_{u\in V}  \sum \limits_{w \in V}\widehat{\textbf{A}}^r_{u,w}\sin(\theta\textbf{x}_w)/|V|&= \sum\limits_{u\in V}  \sum \limits_{w \in V}(\textbf{P} \widehat{\textbf{A}} \textbf{P}^{-1})^r_{u,w}\sin(\theta\cdot(\textbf{P}\textbf{x})_w)/|V|.
\end{align*}

\end{proof}
\subsection{Parametric characteristic functions}
Our discussion postulated that the evaluation points of the $r$-scale random walk characteristic function are predetermined. However, we can define parametric models where these evaluation points are learned in a semi-supervised fashion to make the evaluation points selected the most discriminative with regards to a downstream classification task. The process which we describe in Algorithm \ref{characteristic_function_algo} can be interpreted as the forward pass of a graph neural network which uses the normalized adjacency matrix and node features as input. This way the evaluation points and the weights of the classification model could be learned jointly in an end-to-end fashion.

\subsubsection{Softmax parametric model} Now we define the classifiers with learned evaluation points, let $\textbf{Y}$ be the $|V|\times C$ one-hot encoded matrix of node labels, where $C$ is the number of node classes. Let us assume that $\textbf{Z}$ was calculated by using Algorithm \ref{characteristic_function_algo} in a forward pass with a trainable $\widetilde{\Theta}$. 
The classification weights of the softmax characteristic function classifier are defined by the trainable weight matrix $\beta \in \mathbb{R}^{(2\cdot k\cdot d\cdot r) \times C}$. The class label distribution matrix of nodes outputted by the softmax characteristic function model is defined by Equation \eqref{eq:model_1} where the softmax function is applied row-wise. We reference this supervised softmax model as \textit{FEATHER-L}.
\begin{align}
\widehat{\textbf{Y}}=\text{softmax}(\textbf{Z}\cdot \beta)\label{eq:model_1}
\end{align}

\subsubsection{Neural parametric model} We introduce the forward pass of a neural characteristic function model with a single hidden layer of feed forward neurons. The trainable input weight matrix is $\beta_0 \in \mathbb{R}^{(2\cdot k\cdot d\cdot r) \times h}$ and the output weight matrix is $\beta_1\in \mathbb{R}^{h \times C}$, where $h$ is the number of neurons in the hidden layer. The class label distribution matrix of nodes output by the neural model is defined by Equation \eqref{eq:model_2}, where $\sigma(\cdot)$ is an activation function applied element-wise (in our experiments it is a ReLU). We refer to neural models with this architecture as \textit{FEATHER-N}.
\begin{align}
\widehat{\textbf{Y}}=\text{softmax}( \sigma(\textbf{Z}\cdot\beta_0)\cdot \beta_1)\label{eq:model_2}
\end{align}

\subsubsection{Optimizing the parametric models.} The log-loss of the \textit{FEATHER-N} and \textit{FEATHER-L} models being minimized is defined by Equation \eqref{eq:linear_loss} where $U\subseteq V$ is the set of labeled training nodes.
\begin{align}
    \mathcal{L}&=-\sum\limits_{u\in U}\sum\limits_{c=1}^{C}\textbf{Y}_{u,c}\cdot \text{log}(\widehat{\textbf{Y}}_{u,c})\label{eq:linear_loss}
\end{align}

This loss is minimized with a variant of gradient descent to find the optimal values of $\beta$ (respectively $\beta_0$ and $\beta_1$) and $\widetilde{\Theta}$. The softmax model has $\mathcal{O}(k\cdot r\cdot  d\cdot C)$ while the neural model has $\mathcal{O}(k\cdot r\cdot  d\cdot h + C\cdot h )$ free trainable parameters, As a comparison, generating the representations upstream with \textit{FEATHER} and learning a logistic regression has $\mathcal{O}(k\cdot r\cdot  d\cdot C)$ trainable parameters.

%% file: algorithms/r_scale_characteristic_function_calculation.tex
\begin{algorithm}[h!]
    
    \vspace{5pt}
    \DontPrintSemicolon
    \SetAlgoLined
    \footnotesize
    \KwData{    $\widehat{\textbf{A}}$ -- Normalized adjacency matrix.\\
    \quad \quad\,\,\,\,\,$\mathcal{X}=\left\{\textbf{x}^1, \dots, \textbf{x}^k\right\}$ -- Set of node feature vectors.\\
    \quad \quad\,\,\,\,\,$\widetilde{\Theta}=\left\{\Theta^{1,1}, \dots, \Theta^{1,r},\Theta^{2,1}, \dots, \Theta^{k,r}\right\}$ -- Set of evaluation point vectors.\\    
    \quad \quad\,\,\,\,\,$r$ -- Scale of empirical graph characteristic function.\\
    }
    \vspace{5pt}
    \KwResult{Node embedding matrix $\textbf{Z}$.}
    \vspace{5pt}

$\textbf{Z}_{Re}\leftarrow \text{Initialize Real Features}()$\;  

$\textbf{Z}_{Im}\leftarrow \text{Initialize Imaginary Features}()$\;  

    \For{i\,\,\text{\upshape in}\,\,$1:k$}{

        \For{$j \,\, \textup{in} \,\, 1:r$}{
        \For{$l \,\, \textup{in} \,\, 1:j$}{
\uIf{$l=1$}{
$\textbf{H}\leftarrow \textbf{x}^i \otimes \Theta^{i,j}$\;

$\textbf{H}_{Re} \leftarrow \cos(\textbf{H})$ \;

$\textbf{H}_{Im} \leftarrow \sin(\textbf{H})$\;

}
$\textbf{H}_{Re} \leftarrow \widehat{\textbf{A}}\textbf{H}_{Re} $\;

$\textbf{H}_{Im} \leftarrow \widehat{\textbf{A}}\textbf{H}_{Im}$\;
	  }
$\textbf{Z}_{Re} \leftarrow \left[\textbf{Z}_{Re} \mid \textbf{H}_{Re} \right]$\;

$\textbf{Z}_{Im} \leftarrow \left[\textbf{Z}_{Im} \mid \textbf{H}_{Im} \right]$\;
  
    	}
    }
$\textbf{Z} \leftarrow \left[\textbf{Z}_{Im} \mid \textbf{Z}_{Re} \right]$\;
    
    Output $\textbf{Z}$.  
    {\small\caption{Efficient $r$-scale random walk weighted characteristic function calculation for multiple node features.}\label{characteristic_function_algo}}
    
\end{algorithm}

%% file: sections/experiments.tex
\section{Experimental Evaluation}\label{sec:feather_experiments}
In this section, we overview the datasets used to quantify representation quality. We demonstrate how node and graph features distilled with \textit{FEATHER} can be used to solve node and graph classification tasks. Furthermore, we highlight the transfer learning capabilities, scalability and robustness of our method.  

\subsection{Datasets}
We briefly discuss the real world datasets and their descriptive statistics which we use to evaluate the node and graph features extracted with our proposed methods.
\input{tables/node_level_datasets.tex}
\subsubsection{Node level datasets.} We used various publicly available, and self-collected social network and webgraph datasets to evaluate the quality of node features extracted with \textit{FEATHER}. The descriptive statistics of these datasets are summarized in Table \ref{fig:node_level_statistics}. These graphs are heterogeneous with respect to size, density, and number of features, and they allow for binary and multi-class node classification. 
\begin{itemize} 
\item \textbf{Wikipedia Crocodiles \cite{karateclub}:} A webgraph of Wikipedia articles about crocodiles where each node is a page and edges are mutual links between edges. Attributes represent nouns appearing in the articles and the binary classification task on the dataset is deciding whether a page is popular or not.
\item \textbf{Twitch Social Networks \cite{musae}:} Social networks of gamers from the streaming service Twitch. Features describe the history of games played and the task is to predict whether a gamer streams adult content. The country specific graphs share the same node features which means that we can perform transfer learning with these datasets. 
\item \textbf{Facebook Page-Page \cite{musae}:} A webgraph of verified Facebook pages which liked each other. Features were extracted from page descriptions and the classification target is the page category.
\item \textbf{LastFM Asia:} The LastFM Asia graph is a social network of users from Asian (e.g. Philippines, Malaysia, Singapore) countries which we collected. Nodes represent users of the music streaming service LastFM and links among them are friendships. We collected these datasets in March 2020 via the use of the API. The classification task related to these two datasets is to predict the home country of a user given the social network and artists liked by the user. 
\item \textbf{Deezer Europe:} A social network of European Deezer users which we collected from the public API in March 2020. Nodes represent users and links are mutual follower relationships among users. The related classification task is the prediction of gender using the friendship graph and the artists liked. 
\end{itemize}

\subsubsection{Graph level datasets.} We utilized a range of publicly available non-attributed, social graph datasets to assess the quality of graph level features distilled via our procedure. Summary statistics, enlisted in Table \ref{fig:graph_level_statistics}, demonstrate that these datasets have a large number of small graphs with varying size, density and diameter.
\begin{itemize}
    \item \textbf{Reddit Threads \cite{karateclub}}: A collection of Reddit thread and non-thread based discussion graphs. The related task is to correctly classify graphs according the thread -- non-thread categorization.
    \item \textbf{Twitch Egos \cite{karateclub}} The ego-networks of Twitch users who participated in the partnership program. The classification task entails the identification of gamers who only play with a single game.
    \item \textbf{GitHub Repos \cite{karateclub}:} Social networks of developers who starred machine learning and web design repositories. The target is the type of the repository itself.
    \item \textbf{Deezer Egos \cite{karateclub}:} A small collection of ego-networks for European Deezer users. The related task involves the prediction of the ego node's gender.  
\end{itemize}
\input{tables/graph_level_datasets.tex}

\subsection{Node classification}\label{subsec:node_classification}
The node classification performance of \textit{FEATHER} variants is compared to neighbourhood based, structural and attributed node embedding techniques. We also studied the performance in contrast to various competitive graph neural network architectures.

\subsubsection{Experimental settings}

 We report mean micro averaged test AUC scores with standard errors calculated from 10 seeded splits with a 20\%/80\% train-test split ratio in Table \ref{fig:node_classification_performance}.

The unsupervised neighbourhood based \cite{deepwalk, line, walklets, netmf, diff2vec,grarep}, structural \cite{role2vec} and attributed node \cite{asne, bane, tene, tadw, sine, musae} embeddings were created by the \textit{Karate Club} \cite{karateclub} software package and used the default hyperparameter settings of the 1.0 release. This ensure that the number of free parameters used to represent the nodes by the upstream unsupervised models is the same. We used the publicly available official Python implementation of \textit{Node2Vec} \cite{node2vec} with the default settings and the \textit{In-Out} and \textit{Return} hyperparameters were fine tuned with 5-fold cross validation within the training set. The downstream classifier was a logistic regression which used the default hyperparameter settings of \textit{Scikit-Learn} \cite{scikit} with the \textit{SAGA} optimizer \cite{saga}.

\input{./tables/small_node.tex}

Supervised baselines were implemented with the \textit{PyTorch Geometric} framework \cite{pytorch_geometric} and as a pre-processing step, the dimensionality of vertex features was reduced to be 128 by the \textit{Scikit-Learn} implementation of Truncated SVD \cite{truncatedsvd}. Each supervised model considers neighbours from 2 hop neighbourhoods except \textit{APPNP} \cite{appnp} which used a teleport probability of 0.2 and 10 personalized PageRank approximation iterations. Models were trained with the Adam optimizer \cite{adam} with a learning rate of 0.01 for 200 epochs. The input layers of the models had 32 filters and between the final and input layer we used a 0.5 dropout rate \cite{dropout} during training time. The \textit{ClusterGCN} \cite{clustergcn} models decomposed the graph with the \textit{METIS} \cite{metis} algorithm before training -- the number of clusters equaled the number of classes.

The \textit{FEATHER} model variants used a combination of neighbourhood based, structural and generic vertex attributes as input besides the graph itself. Specifically we used:
\begin{itemize}
    \item \textbf{Neighbourhood features:} We used Truncated SVD to extract 32 dimensional node features from the normalized adjacency matrix.
    \item \textbf{Structural features:} The log transformed degree and the clustering coefficient are used as structural vertex features.
    \item \textbf{Generic node features:} We reduced the dimensionality of generic vertex features with Truncated SVD to be 32 dimensional for each network.
\end{itemize}

The unupservised \textit{FEATHER} model used 16 evaluation points per feature, which were initialized uniformly in the $\left[0,5\right]$ domain, and a scale of $r=2$. We used a logistic regression downstream classifier. The characteristic function evaluation points of the supervised \textit{FEATHER-L} and \textit{FEATHER-N} models were initialized similarly. Models were trained by the Adam optimizer with a learning rate of 0.001, for 50 epochs and the neural model had 32 neurons in the hidden layer. 

\subsubsection{Node classification performance} Our results in Table \ref{fig:node_classification_performance} demonstrate that the unsupervised \textit{FEATHER} algorithm outperforms the proximity preserving, structural and attributed node embedding techniques. This predictive performance advantage varies between 0.4\% and 4.6\% in terms of micro averaged test AUC score. On the Wikipedia, Facebook and LastFM datasets, the performance difference is significant at an $\alpha=1\%$ significance level. On these three datasets the best supervised \textit{FEATHER} variant marginally outperforms  \textit{graph neural networks} between 0.1\% and 2.1\% in terms of test AUC. However, this improvement of classification is only significant on the Wikipedia dataset at $\alpha=1\%$. 

\subsection{Graph classification}
The graph classification performance of unsupervised and supervised \textit{FEATHER} models is compared to that of implicit matrix factorization, spectral fingerprinting and graph neural network models.
\subsubsection{Experimental settings}
We report average test AUC values with standard errors on binary classification tasks calculated from 10 seeded splits with a 20\%/80\% train-test split ratio in Table \ref{fig:graph_classification_performance}.

The unsupervised implicit matrix factorization \cite{graph2vec, gl2vec} and spectral fingerprinting \cite{sf, fgsd, netlsd, geoscattering} representations were produced by the \textit{Karate Club} framework with the standard hyperparameter settings of the 1.0 release. The downstream graph classifier was a logistic regression model implemented in \textit{Scikit-Learn}, with the standard hyperparameters and the \textit{SAGA} \cite{saga} optimizer.

Supervised models used the one-hot encoded degree, clustering coefficient and eccentricity as node features. Each method was trained by minimizing log-loss with the Adam optimizer \cite{adam} using a learning rate of 0.01 for 10 epochs with a batch size of 32. We used two consecutive graph convolutional \cite{gcn} layers with 32 and 16 filters and ReLu activations. In the case of mean and maximum pooling the output of the second convolutional layer was pooled and fed to a fully connected layer. We do the same with Sort Pooling \cite{sortpooling} by keeping 5 nodes and flattening the output of the pooling layer to create graph representations. In the case of Top-K pooling \cite{topkpooling} and SAG Pooling \cite{sagpooling} we pool the nodes after each convolutional layer with a pooling ratio of 0.5 and output graph representations with a final max pooling layer. The output of advanced pooling layers \cite{sortpooling, topkpooling, sagpooling} was fed to a fully connected layer. The output of the final layers was transformed by the softmax function.

We used the unsupervised \textit{FEATHER} model to create graph descriptors. We pooled the node features extracted for each characteristic function evaluation point with a permutation invariant aggregation function such as the mean, maximum and minimum. Node level representations only used the log transformed degree  as a feature. We set $r=5$, $d=25$, and initialized the characteristic function evaluation points in the $[0,5]$ interval uniformly. Using these descriptors we utilized logistic regression as a classifier with the settings used with other unsupervised methods.

\subsubsection{Graph classification performance}\label{subsec:graph_classification} 

Our results demonstrate that our proposed pooled characteristic function based classification method outperforms both supervised and unsupervised graph classification methods on the Reddit Threads, Twitch Egos and Github Repos datasets. The performance advantage of \textit{FEATHER} varies between 1.1\% and 12.0\% in terms of AUC, which is a significant peformance gain on all three of these datasets at an $\alpha=1\%$ significance level. On the Deezer Egos dataset the disadvantage of our method is not significant, but specific supervised and unsupervised procedures have a somewhat better predictive performance in terms of test AUC. We also have evidence that the mean pooling of the node level characteristic functions provides superior peformance on most datasets considered.

\input{tables/graph_classification_table}

\subsection{Sensitivity analysis}
We investigated how the representation quality changes when the most important hyperparameters of the $r$-scale random walk weighted characteristic function are manipulated. Precisely, we looked  at the scale of the $r$-scale random walk weighted characteristic function and the number of evaluation points.

We use the Facebook Page-Page dataset, with the standard (20\% /80\%) split and input the log transformed degree as a vertex feature. Figure \ref{fig:feature_dimension_sensitivity} plots the average test AUC against the manipulated hyperparameter calculated from 10 seeded splits. The models were trained with the hyperparameter settings discussed in Subsection \ref{subsec:node_classification}. We chose a scale of 5 when the number of evaluation points was modulated and used 25 evaluation points when the scale was manipulated. The evaluation points were initialized uniformly in the $[0,5]$ domain.

\input{figures/feature_dimensions_sensitivity}

\subsubsection{Scale of the characteristic function.} First, we observe that including information from higher order neighbourhoods is valuable for the classification task. Second, the marginal performance increase is decreasing with the increase of the scale. Finally, when we only consider the first hop of neighbours we observe little performance difference between the unsupervised and supervised model variants. When higher order neighbourhoods are considered the supervised models have a considerable advantage.  

\subsubsection{Number of evaluation points.} Increasing the number of characteristic function evaluation points increases the performance on the downstream predictive task. Supervised models are more efficient when the number of characteristic function evaluation points was low. The neural model is efficient in terms of the evaluation points needed for a good predictive performance. It is evident that the marginal predictive performance gains of the supervised models are decreasing with the number of evaluation points.

\subsection{Transfer learning}\label{subsec:transfer_learning}
Using the Twitch datasets, we demonstrate that the $r$-scale random walk weighted characteristic function features are robust and can be easily utilized in a transfer learning setup. We support evidence that the supervised parametric models also work in such scenarios. Figure \ref{char_transfer} shows the transfer learning results for German, Spanish and Portuguese users, where the predictive performance is measured by average AUC values based on 10 experiments.

Each model was trained with nodes of a fully labeled source graph and evaluated on the nodes of the target graph. This transfer learning experiment requires that graphs share the target variable (abusing the Twitch platform), and that the characteristic function is calculated for the same set of node features. All models utilized the log transformed degree centrality of nodes as a shared and cheap-to-calculate structural feature. We used a scale of $r=5$ and $d=25$ characteristic function evaluation points for each \textit{FEATHER} model variant. Models were fitted with the hyperparameter settings described in Subsection  \ref{subsec:node_classification}.

\input{./figures/node_transfer_linear.tex}

Firstly, our results presented on Figure \ref{char_transfer}
 support that even the characteristic function of a single structural feature is sufficient for transfer learning as we are able to outperform the random guessing of labels in most transfer scenarios. Secondly, we see that the neural model has a predictive performance advantage over the unsupervised \textit{FEATHER} and the shallow \textit{FEATHER-L} model. Specifically, for the Portuguese users this advantage varies between 2.1 and 3.3\% in terms of average AUC value. Finally, transfer from the English users seems to be poor to the other datasets. Which implies that the abuse of the platform is associated with different structural features in that case. 

\subsection{Runtime performances}
We evaluated the runtime needed for calculating the proposed $r$-scale random walk weighted characteristic function. Using a synthetic Erdős-Rényi graph with $2^{12}$nodes and $2^4$ edges per node, we measured the runtime of Algorithm \ref{characteristic_function_algo} for various values of the size of the graph, number of features and characteristic function evaluation points. Figure \ref{fig:feather_runtime} shows the mean logarithm of the runtimes against the manipulated input parameter based on 10 experimental repetitions.

\input{./figures/runtimes.tex}

Our results support the theoretical runtime complexities discussed in Subsection \ref{subsec:runtime}. Practically it means that doubling the number of nodes, edges, features or characteristic function evaluation points doubles the expected runtime of the algorithm. Increasing the scale of random walks (considering more hops) increases the runtime. However, for small values of $r$ the quadratic increase in runtime is not evident.

%% file: tables/node_level_datasets.tex
\begin{table}[htbp]
	\centering
	\vspace{-2mm}
	\caption{Statistics of social networks used for the evaluation of node classification algorithms, sensitivity analysis, and transfer learning.}
	\label{fig:node_level_statistics}
	
	{\footnotesize
		
	\setlength\tabcolsep{3pt}	\begin{tabular}{ccccccccc}
			\hline
			\textbf{Dataset}& \textbf{Nodes} & $\textbf{Density}$ & \specialcell{\textbf{Clustering}\\\textbf{Coefficient}}& \textbf{Diameter}& \specialcell{\textbf{Unique}\\\textbf{Features}}&\textbf{Classes}\\ \hline
			\textbf{Wiki Croco}& 11,631 & 0.003 & 0.026& 11&13,183&2\\ [0.3em]
			\textbf{FB Page-Page}& 22,470 & 0.001 &0.232& 15 &4,714&4\\ [0.3em]

			\textbf{LastFM ASIA}& 7,624 & 0.001 & 0.179& 15&7,842&18\\ [0.3em]
			\textbf{Deezer EUR}& 28,281 & 0.002 & 0.096& 21&31,240&2\\ [0.3em]
			\hline
			\textbf{Twitch DE}& 9,498 & 0.003 & 0.047& 7&3,169&2\\ [0.3em]
			\textbf{Twitch EN}& 7,126 & 0.002 & 0.042& 10&3,169&2\\ [0.3em]
			\textbf{Twitch ES}& 4,648 & 0.006 & 0.084& 9&3,169&2\\ [0.3em]
			\textbf{Twitch PT}& 1,912 & 0.017 & 0.131& 7&3,169&2\\ [0.3em]
			\textbf{Twitch RU}& 4,385 & 0.004 & 0.049& 9&3,169&2\\ [0.3em]
			\textbf{Twitch TW}& 2,772 & 0.017 &0.120& 7&3,169&2\\ [0.3em]
			\hline
		\end{tabular}}
			\vspace{-4mm}		
	\end{table}

%% file: tables/graph_level_datasets.tex
\begin{table}[htbp]
	\centering
			\vspace{-2mm}
	\caption{Statistics of graph datasets used for the evaluation of graph classification algorithms.}
	\label{fig:graph_level_statistics}
	
	{\footnotesize
\setlength\tabcolsep{4pt} 

\begin{tabular}{cccccccc}
            &        & \multicolumn{2}{c}{\textbf{Nodes}} & \multicolumn{2}{c}{\textbf{Density}} & \multicolumn{2}{c}{\textbf{Diameter}} \\[0.25em]
            \cline{3-8} 
\textbf{Dataset}     & \textbf{Graphs} & \textbf{Min}         & \textbf{Max}         & \textbf{Min}          & \textbf{Max}          & \textbf{Min}           & \textbf{Max}          \\[0.25em]\hline
\textbf{Reddit Threads} &   203,088     &      11      &         97    & 0.021             &  0.382     & 2               &   27           \\[0.3em]
\textbf{Twitch Egos}   &127,094&14 &52 &0.038  &0.967  &  1             &       2       \\[0.3em]
\textbf{GitHub Repos}   & 12,725      &    10        &957      & 0.003     &       0.561            &     2          &18              \\[0.3em]
\textbf{Deezer Egos}   &   9,629     &   11          &     363        &           0.015   &    0.909          &    2           &      2        \\[0.3em]
			\hline
		\end{tabular}}
			\vspace{-4mm}
	\end{table}

%% file: tables/small_node.tex
\begin{table}[ht!]
	\vspace{-3pt}
	\caption[Mean micro-averaged AUC values on the test set with standard errors on the node level datasets calculated from 10 seed train-test splits. Black bold numbers denote the best performing unsupervised model while blue ones denote the best supervised one.]{Mean micro-averaged AUC values on the test set with standard errors on the node level datasets calculated from 10 seed train-test splits. Black bold numbers denote the best performing unsupervised model, while blue ones denote the best supervised one.}
	\label{fig:node_classification_performance}
		{\small
	\centering{
\setlength\tabcolsep{3pt}		\begin{tabular}{lcccc}
			& \specialcell{\textbf{Wikipedia}\\\textbf{Crocodiles}} & \specialcell{\textbf{Facebook}\\\textbf{Page-Page}}&  \specialcell{\textbf{LastFM}\\\textbf{Asia}}& \specialcell{\textbf{Deezer}\\\textbf{Europe}} \\[0.45em]
			\hline
			\textbf{DeepWalk} \cite{deepwalk}       &$.820\pm.001$&$.880\pm.001$&$.918\pm.001$&$.520\pm.001$\\[0.35em]
			\textbf{LINE}  \cite{line}          &$.856\pm.001$&$.956\pm.001$&$.949\pm.001$&$.543\pm.001$\\[0.35em]
			\textbf{Walklets} \cite{walklets}          &$.872\pm.001$&$.975\pm.001$&$\textbf{.950}\pm\textbf{.001}$&$.547\pm.001$\\[0.35em]
			\textbf{HOPE} \cite{hope}            &$.855\pm.001$&$.903\pm.002$&$.922\pm.001$&$.539\pm.001$\\[0.35em]
			\textbf{NetMF}  \cite{netmf}          &$.859\pm.001$&$.946\pm.001$&$.943\pm.001$&$.538\pm.001$\\[0.35em]
			\textbf{Node2Vec} \cite{node2vec}           &$.850\pm.001$&$.974\pm.001$&$.944\pm.001$&$.556\pm.001$\\[0.35em]
	\textbf{Diff2Vec} \cite{diff2vec}         &$.812\pm.001$&$.867\pm.001$&$.907\pm.001$&$.521\pm.001$\\[0.35em]
	\textbf{GraRep} \cite{grarep}           &$.871\pm.001$&$.951\pm.001$&$.926\pm.001$&$.547\pm.001$\\[0.35em]
	\textbf{Role2Vec} \cite{role2vec}         &$.801\pm.001$&$.911\pm.001$&$.924\pm.001$&$.534\pm.001$\\[0.35em]
	\textbf{GEMSEC} \cite{gemsec}         &$.858\pm.001$&$.933\pm.001$&$\textbf{.951}\pm\textbf{.001}$&$.544\pm.001$\\[0.35em]
	\hline
			\textbf{ASNE} \cite{asne}      &$.853\pm.001$&$.933\pm.001$&$.910\pm.001$&$.528\pm.001$\\[0.35em]
			\textbf{BANE} \cite{bane}        &$.534\pm.001$&$.866\pm.001$&$.610\pm.001$&$.521\pm.001$\\[0.35em]
			\textbf{TENE} \cite{tene}          &$.893\pm.001$&$.874\pm.001$&$.855\pm.002$&$.639\pm.002$\\[0.35em]
			\textbf{TADW} \cite{tadw}        &$.901\pm.001$&$.849\pm.001$&$.851\pm.001$&$.644\pm.001$\\[0.35em]
			\textbf{SINE} \cite{sine}       &$.895\pm.001$&$.975\pm.001$&$.944\pm.001$&$.618\pm.001$\\[0.35em]

			\hline
			
				\textbf{GCN}  \cite{gcn}          &$.924\pm.001$&${\color{blue}\textbf{.984}\pm\textbf{.001}}$&$.962\pm.001$&$.632\pm.001$\\[0.35em]
				\textbf{GAT} \cite{gat}         &$.917\pm.002$&${\color{blue}\textbf{.984}\pm\textbf{.001}}$&$.956\pm.001$&$.611\pm.002$\\[0.35em]
				\textbf{GraphSAGE} \cite{graphsage}          &$.916\pm.001$&${\color{blue}\textbf{.984}\pm.\textbf{001}}$&$.955\pm.001$&$.618\pm.001$\\[0.35em]
				\textbf{ClusterGCN} \cite{clustergcn}           &$.922\pm.001$&$.977\pm.001$&$.944\pm.002$&$.594\pm.002$\\[0.35em]
				\textbf{APPNP} \cite{appnp}       &$.900\pm.001$&${\color{blue}\textbf{.986}\pm\textbf{.001}}$&{\color{blue}$\textbf{.968}\pm\textbf{.001}$}&$.667\pm.001$\\[0.35em]
				\textbf{MixHop} \cite{mixhop}           &$.928\pm.001$&$.976\pm.001$&$.956\pm.001$&{\color{blue}$\textbf{.682}\pm\textbf{.001}$}\\[0.35em]
				\textbf{SGConv} \cite{sgconv}           &$.889\pm.001$&$.966\pm.001$&$.957\pm.001$&$.647\pm.001$\\[0.35em]
				\hline
			\textbf{FEATHER}            &$\textbf{.943}\pm\textbf{.001}$&$\textbf{.981}\pm\textbf{.001}$&$\textbf{.954}\pm\textbf{.001}$&$\textbf{.651}\pm\textbf{.001}$\\[0.35em]
			\textbf{FEATHER-L}            &{\color{blue}$\textbf{.944}\pm\textbf{.002}$}&${\color{blue}\textbf{.984}\pm\textbf{.001}}$&$.960\pm.001$&$.656\pm.001$\\[0.35em]
\textbf{FEATHER-N}            &{\color{blue}$\textbf{.947}\pm\textbf{.001}$}&${\color{blue}\textbf{.987}\pm\textbf{.001}}$&{\color{blue}$\textbf{.970}\pm\textbf{.001}$}&$.673\pm.001$\\[0.35em]
			\hline \\
	
	\end{tabular}}}
	\vspace{-6mm}
	\end{table}

%% file: tables/graph_classification_table.tex
\begin{table}[h!]
	
	\caption[Mean AUC values with standard errors on the graph datasets calculated from 10 seed train-test splits. Bold numbers denote the model with the best performance.]{Mean AUC values with standard errors on the graph datasets calculated from 10 seed train-test splits. Bold numbers denote the model with the best performance.}
	\label{fig:graph_classification_performance}
		{\small
		\setlength\tabcolsep{4pt}
	\centering{
		\begin{tabular}{lcccc}
			& \specialcell{\textbf{Reddit}\\\textbf{Threads}} &\specialcell{ \textbf{Twitch}\\\textbf{Egos}} & \specialcell{\textbf{GitHub}\\\textbf{Repos}}&  \specialcell{\textbf{Deezer}\\\textbf{Egos}} \\[0.45em]
			\hline
			\textbf{GL2Vec} \cite{gl2vec}         &$.754\pm.001$&$.670\pm.001$&$.532\pm.002$&$.500\pm.001$\\[0.35em]
			\textbf{Graph2Vec}  \cite{graph2vec}          &$.808\pm.001$&$.698\pm.001$&$.563\pm.002$&$.510\pm.001$\\[0.35em]
			\textbf{SF} \cite{sf}           &$.819\pm.001$&$.642\pm.001$&$.535\pm.001$&$.503\pm.001$\\[0.35em]
			\textbf{NetLSD} \cite{netlsd}           &$.817\pm.001$&$.630\pm.001$&$.614\pm.002$&$.525\pm.001$\\[0.35em]
			\textbf{FGSD} \cite{fgsd}           &$.822\pm.001$&$.699\pm.001$&$.650\pm.002$&$\textbf{.528}\pm\textbf{.001}$\\[0.35em]
			\textbf{Geo-Scatter} \cite{geoscattering}           &$.800\pm.001$&$.695\pm.001$&$.532\pm.001$&$.524\pm.001$\\[0.35em]
			\hline
			\textbf{Mean Pool}            &$.801\pm.002$&$.708\pm.001$&$.599\pm.003$&$.503\pm.001$\\[0.35em]
			\textbf{Max Pool}        &$.805\pm.001$&$.713\pm.001$&$.612\pm.013$&$.515\pm.001$\\[0.35em]
			\textbf{Sort Pool} \cite{sortpooling}            &$.807\pm.001$&$.712\pm.001$&$.614\pm.010$&$\textbf{.528}\pm\textbf{.001}$\\[0.35em]
			\textbf{Top K Pool}   \cite{topkpooling}         &$.807\pm.001$&$.706\pm.002$&$.634\pm.001$&$.520\pm.003$\\[0.35em]
			\textbf{SAG Pool}            \cite{sagpooling} &$.804\pm.001$&$.705\pm.002$&$.620\pm.001$&$.518\pm.003$\\[0.35em]
			\hline
			\textbf{FEATHER MIN}            &$\textbf{.834}\pm\textbf{.001}$&$\textbf{.719}\pm\textbf{.001}$&$.694\pm.001$&$.518\pm.001$\\[0.35em]
			\textbf{FEATHER MAX}            &$\textbf{.831}\pm\textbf{.001}$&$\textbf{.718}\pm\textbf{.001}$&$.689\pm.001$&$.521\pm.002$\\[0.35em]
			\textbf{FEATHER AVE}            &$.823\pm.001$&$\textbf{.719}\pm\textbf{.001}$&$\textbf{.728}\pm\textbf{.002}$&$\textbf{.526}\pm\textbf{.001}$\\[0.35em]
			\hline \\
		\end{tabular}}
		}
		\vspace{-8mm}
	\end{table}

%% file: figures/feature_dimensions_sensitivity.tex
\begin{figure}[h!]
	\centering
	\begin{tikzpicture}[scale=0.4,transform shape]
	\tikzset{font={\fontsize{18pt}{12}\selectfont}}
	\begin{groupplot}[group style={group size=2 by 1,
		horizontal sep=60pt, vertical sep=60pt,ylabels at=edge left},
	width=0.58\textwidth,
	height=0.4\textwidth,
	grid=major,
	ymin=0.7,
	ymax=0.95,
	grid style={dashed, gray!40},
	scaled ticks=false,
	inner axis line style={-stealth}]
	
		\nextgroupplot[ytick={0.5,0.6,0.7,0.8,0.9,1.0},
	xtick={1,2,3,4,5},
	xlabel=Random walk scale $(r)$,
	ylabel=Area under the curve,
	enlargelimits=0.1,
	legend style = { column sep = 10pt, legend columns = -1, legend to name = grouplegend,}]
	
	\addplot[mark=triangle*,opacity=0.8,mark options={black,fill=red},mark size=5pt]
	coordinates {

(1,.738)
(2,.821)
(3,.861)
(4,.882)
(5,.893)

	};\addlegendentry{Unsupervised}%
	\addplot[mark=diamond*,opacity=0.8,mark options={black,fill=blue},mark size=5pt]
	coordinates {
(1,.750)
(2,.836)
(3,.887)
(4,.916)
(5,.932)

	};\addlegendentry{Linear}%
	\addplot[mark=*,opacity=0.8,mark options={black,fill=green},mark size=3pt]
	coordinates {
(1,.762)
(2,.849)
(3,.896)
(4,.917)
(5,.934)
		
	};\addlegendentry{Neural}%

	\nextgroupplot[ytick={0.8,0.85,0.9,0.95},
	xtick={5,10,15,20,25,30},
	xlabel=Evaluation points ($d$),
	ymin=0.8,
	ymax=0.95,	
	enlargelimits=0.1,
	legend style = { column sep = 10pt, legend columns = -1, legend to name = grouplegend,}]
	
	\addplot[mark=triangle*,opacity=0.8,mark options={black,fill=red},mark size=5pt]
	coordinates {
(5,0.816)
(10,0.835)
(15,0.856)
(20,0.885)
(25,0.893)
(30,0.912)

	};\addlegendentry{\textbf{FEATHER}}%
	\addplot[mark=diamond*,opacity=0.8,mark options={black,fill=blue},mark size=5pt]
	coordinates {
(5,.841)
(10,.890)
(15,.911)
(20,.922)
(25,.933)
(30,.936)

	};\addlegendentry{\textbf{FEATHER-L}}%
	\addplot[mark=*,opacity=0.8,mark options={black,fill=green},mark size=3pt]
	coordinates {
(5,.898)
(10,.918)
(15,.926)
(20,.933)
(25,.934)
(30,.937)		
	};\addlegendentry{\textbf{FEATHER-N}}%

	\end{groupplot}
	\node at ($(group c1r1) + (6.0cm,-4.7cm)$) {\ref{grouplegend}}; 
	\end{tikzpicture}
    \caption{Mean AUC values on the Facebook page-page test set (10 seeded splits) achieved by FEATHER model variants as a function of random walk scale and characteristic function evaluation point count.}\label{fig:feature_dimension_sensitivity}
	\vspace{-6pt}
\end{figure}
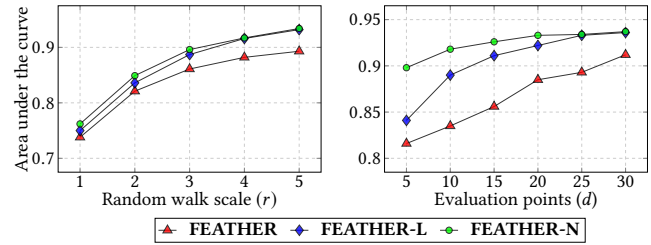

%% file: figures/node_transfer_linear.tex
\begin{figure*}[ht!]
	\centering
	\begin{tikzpicture}[scale=0.45,transform shape]
	\tikzset{font={\fontsize{12pt}{12}\selectfont}}
	\begin{groupplot}[group style={group size=3 by 1,
		horizontal sep=40pt, vertical sep=70pt,ylabels at=edge left},
	width=0.54\textwidth,
	height=0.3375\textwidth,
	ymin=0.5,
	ymax=0.7,
every tick label/.append style={font=\bf},
    y tick label style={
        /pgf/number format/.cd,
            fixed,
            fixed zerofill,
            precision=2,
        /tikz/.cd
    },
 enlarge x limits=true,
	grid=major,
	title=AE,	
	grid style={dashed, gray!40},
	scaled ticks=false,
	inner axis line style={-stealth}]

 \nextgroupplot[
   title=\textbf{Transfer to DE users},
    ybar=0pt,
      every tick/.style={
        black,
        semithick,
      },
legend image post style={solid},     
    bar width=9pt,
    enlargelimits=0.15,
    legend style={at={(0.5,-0.15)},
      anchor=north,legend columns=-1},
    ylabel=\textbf{{Area Under the Curve}},
    ytick={.5,0.55,0.6,0.65,0.7},
    symbolic x coords={EN,ES,PT,RU,TW},
    xtick={EN,ES,PT,RU,TW}]

\coordinate (BASE) at (axis cs:EN,.5);
\coordinate (O1) at (rel axis cs:0,0);
\coordinate (O2) at (rel axis cs:1,0);

\draw [black,ultra thick,sharp plot,opacity=0.4] (BASE -| O1) -- (BASE -| O2);

\coordinate (SKETCH) at (axis cs:EN,.684);
\coordinate (O1) at (rel axis cs:0,0);
\coordinate (O2) at (rel axis cs:1,0);


\coordinate (LINEAR) at (axis cs:EN,.6);
\coordinate (O1) at (rel axis cs:0,0);
\coordinate (O2) at (rel axis cs:1,0);


\coordinate (NEURAL) at (axis cs:EN,.7);
\coordinate (O1) at (rel axis cs:0,0);
\coordinate (O2) at (rel axis cs:1,0);


\addplot [fill=blue!45]  coordinates {(EN,0.518) (ES,0.604) (PT,0.659) (RU,0.633) (TW,0.603)};
\addplot [fill=red!45]  coordinates {(EN,0.476) (ES,0.629) (PT,0.655) (RU,0.607) (TW,0.612)};
\addplot [fill=green!45] coordinates {(EN,0.544) (ES,0.656) (PT,0.668) (RU,0.611) (TW,0.631)};

 \nextgroupplot[
   title=\textbf{Transfer to ES users},
    ybar=0pt,
      every tick/.style={
        black,
        semithick,
      },
    bar width=9pt,
    enlargelimits=0.15,
    legend style={at={(0.5,-0.15)},
      anchor=north,legend columns=-1},
    ytick={.5,0.55,0.6,0.65,0.7},
    symbolic x coords={DE, EN, PT, RU, TW},
    xtick={DE, EN, PT, RU, TW}]

\coordinate (BASE) at (axis cs:DE,.5);
\coordinate (O1) at (rel axis cs:0,0);
\coordinate (O2) at (rel axis cs:1,0);

\draw [black,ultra thick,sharp plot,opacity=0.4] (BASE -| O1) -- (BASE -| O2);

\coordinate (SKETCH) at (axis cs:DE,.654);
\coordinate (O1) at (rel axis cs:0,0);
\coordinate (O2) at (rel axis cs:1,0);


\coordinate (LINEAR) at (axis cs:DE,.6);
\coordinate (O1) at (rel axis cs:0,0);
\coordinate (O2) at (rel axis cs:1,0);


\coordinate (NEURAL) at (axis cs:DE,.662);
\coordinate (O1) at (rel axis cs:0,0);
\coordinate (O2) at (rel axis cs:1,0);


\addplot [fill=blue!45]  coordinates {(DE,0.631) (EN,0.551) (PT,0.631) (RU,0.602) (TW,0.608)};
\addplot [fill=red!45]  coordinates {(DE,0.635) (EN,0.562) (PT,0.627) (RU,0.602) (TW,0.621)};
\addplot [fill=green!45] coordinates {(DE,0.641) (EN,0.562) (PT,0.642) (RU,0.606) (TW,0.609)};

 \nextgroupplot[
   title=\textbf{Transfer to PT users},
    ybar=0pt,
      every tick/.style={
        black,
        semithick,
      },
    bar width=9pt,
    enlargelimits=0.15,
    legend image post style={solid},
    legend style={at={(0.5,-0.15)},
      anchor=north,legend columns=-1},
    ytick={.5,0.55,0.6,0.65,0.7},
    symbolic x coords={DE, EN, ES, RU, TW},
    xtick={DE, EN, ES, RU, TW},
    	legend style = { column sep = 10pt, legend columns = 1, legend to name = grouplegend}   ]

\coordinate (BASE) at (axis cs:EN,.5);
\coordinate (O1) at (rel axis cs:0,0);
\coordinate (O2) at (rel axis cs:1,0);

\draw [black,ultra thick,solid,opacity=0.4] (BASE -| O1) -- (BASE -| O2);\addlegendentry{Random guessing}\addlegendimage{line legend,solid, color=black, mark options={solid},mark=square*}

\coordinate (SKETCH) at (axis cs:EN,.648);
\coordinate (O1) at (rel axis cs:0,0);
\coordinate (O2) at (rel axis cs:1,0);


\coordinate (LINEAR) at (axis cs:EN,.6);
\coordinate (O1) at (rel axis cs:0,0);
\coordinate (O2) at (rel axis cs:1,0);


\coordinate (NEURAL) at (axis cs:EN,.7);
\coordinate (O1) at (rel axis cs:0,0);
\coordinate (O2) at (rel axis cs:1,0);


\addplot [fill=blue!45]  coordinates {(DE,0.641) (EN,0.543) (ES,0.641) (RU,0.600) (TW,0.598)};\addlegendentry{FEATHER transfer}
\addplot [fill=red!45]  coordinates {(DE,0.642) (EN,0.543) (ES,0.639) (RU,0.606) (TW,0.602)};\addlegendentry{FEATHER-L transfer}
\addplot [fill=green!45] coordinates {(DE,0.660) (EN,0.559) (ES,0.662) (RU,0.619) (TW,0.618)};\addlegendentry{FEATHER-N transfer}
	\end{groupplot}
	\node at ($(group c3r1) + (7.9cm,0.cm)$) {\ref{grouplegend}}; 
	\end{tikzpicture}
	\caption{Transfer learning performance of FEATHER variants on the Twitch Germany, Spain and Portugal datasets as target graphs.   The transfer performance was evaluated by mean AUC values calculated from 10 seeded experimental repetitions.}\label{char_transfer}
\end{figure*}
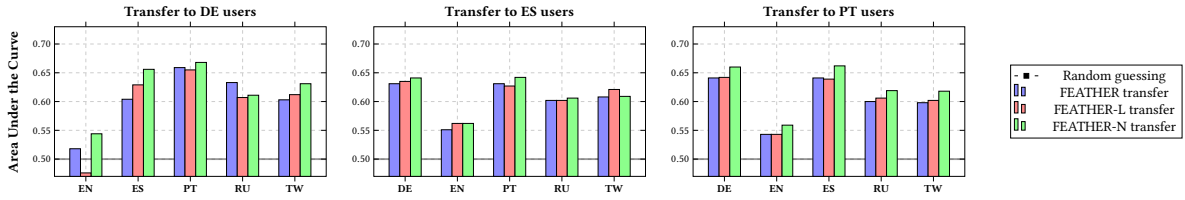

%% file: figures/runtimes.tex
\begin{figure}[h!]
	\centering
	\begin{tikzpicture}[scale=0.4,transform shape]
	\tikzset{font={\fontsize{18pt}{12}\selectfont}}
	\begin{groupplot}[group style={group size=2 by 2,
		horizontal sep=60pt, vertical sep=60pt,ylabels at=edge left},
	width=0.6\textwidth,
	height=0.4\textwidth,
	grid=major,
	grid style={dashed, gray!40},
	scaled ticks=false,
	inner axis line style={-stealth}]

	\nextgroupplot[ytick={-6,-4,-2,0,2,4},
	xtick={10,12,14,16},
	xlabel=$\log_2$ Number of vertices,
	ylabel=$\log_2$ Runtime in seconds,
	enlargelimits=0.1,
	legend style = { column sep = 10pt, legend columns = -1, legend to name = grouplegend,}]
	
	\addplot[mark=triangle*,opacity=0.8,mark options={black,fill=red},mark size=5pt]
	coordinates {

(10,-2.633)
(11,-1.567)
(12,-0.485)
(13,0.561)
(14,1.634)
(15,2.725)
(16,3.84)
	};\addlegendentry{1 core}%
	\addplot[mark=diamond*,opacity=0.8,mark options={black,fill=blue},mark size=5pt]
	coordinates {

(10,-2.399)
(11,-1.314)
(12,-0.238)
(13,0.816)
(14,1.888)
(15,2.947)
(16,4.05)
	};\addlegendentry{2nd order}%
	\addplot[mark=*,opacity=0.8,mark options={black,fill=green},mark size=3pt]
	coordinates {

(10,-2.292)
(11,-1.182)
(12,-0.03)
(13,0.998)
(14,2.081)
(15,3.144)
(16,4.243)

	};\addlegendentry{3rd order}%
	
	\addplot[mark=square*,opacity=0.8,mark options={black,fill=yellow},mark size=3pt]
	coordinates {

(10,-2.118)
(11,-0.969)
(12,0.165)
(13,1.183)
(14,2.268)
(15,3.325)
(16,4.409)

	};\addlegendentry{4th order}%

	
	\nextgroupplot[ytick={-1,1,3,5},
	xtick={3,5,7,9},
	xlabel=$\log_2$ Number of edges per node,
	ylabel=$\log_2$ Runtime in seconds,
	enlargelimits=0.1,
	legend style = { column sep = 10pt, legend columns = -1, legend to name = grouplegend,}]
	
	\addplot[mark=triangle*,opacity=0.8,mark options={black,fill=red},mark size=5pt]
	coordinates {

(3,-1.05)
(4,-0.503)
(5,0.233)
(6,1.121)
(7,2.079)
(8,3.146)
(9,4.244)

	};\addlegendentry{1 core}%
	\addplot[mark=diamond*,opacity=0.8,mark options={black,fill=blue},mark size=5pt]
	coordinates {
(3,-0.922)
(4,-0.307)
(5,0.504)
(6,1.383)
(7,2.401)
(8,3.499)
(9,4.53)

	};\addlegendentry{2nd order}%
	\addplot[mark=*,opacity=0.8,mark options={black,fill=green},mark size=3pt]
	coordinates {
(3,-0.777)
(4,-0.071)
(5,0.763)
(6,1.679)
(7,2.655)
(8,3.749)
(9,4.715)
	};\addlegendentry{3rd order}%
	
	\addplot[mark=square*,opacity=0.8,mark options={black,fill=yellow},mark size=3pt]
	coordinates {
(3,-0.548)
(4,0.117)
(5,0.952)
(6,1.879)
(7,2.836)
(8,3.916)
(9,4.986)

	};\addlegendentry{4th order}%
	
		
	\nextgroupplot[ytick={-6,-4,-2,0,2,4},
	xtick={1,3,5,7},
	xlabel=$\log_2$ Number of features,
	ylabel=$\log_2$ Runtime in seconds,
	enlargelimits=0.1,
	legend style = { column sep = 10pt, legend columns = -1, legend to name = grouplegend,}]
	
	\addplot[mark=triangle*,opacity=0.8,mark options={black,fill=red},mark size=5pt]
	coordinates {
(1,-1.311)
(2,-1.183)
(3,-0.923)
(4,-0.503)
(5,0.135)
(6,0.775)
(7,1.605)
	};\addlegendentry{1st order}%
	\addplot[mark=diamond*,opacity=0.8,mark options={black,fill=blue},mark size=5pt]
	coordinates {
(1,-1.299)
(2,-1.13)
(3,-0.754)
(4,-0.242)
(5,0.514)
(6,1.167)
(7,2.034)
	};\addlegendentry{2nd order}%
	\addplot[mark=*,opacity=0.8,mark options={black,fill=green},mark size=3pt]
	coordinates {
(1,-1.239)
(2,-1.028)
(3,-0.591)
(4,-0.008)
(5,0.794)
(6,1.475)
(7,2.367)
	};\addlegendentry{3rd order}%
	
	\addplot[mark=square*,opacity=0.8,mark options={black,fill=yellow},mark size=3pt]
	coordinates {
(1,-1.195)
(2,-0.955)
(3,-0.502)
(4,0.188)
(5,1.048)
(6,1.735)
(7,2.641)
	};\addlegendentry{4th order}%
		
	\nextgroupplot[ytick={-6,-4,-2,0,2,4},
	xtick={1,3,5,7},
	xlabel=$\log_2$ Number of evaluation points,
	ylabel=$\log_2$ Runtime in seconds,
	enlargelimits=0.1,
	legend style = { column sep = 10pt, legend columns = -1, legend to name = grouplegend,}]
	
	\addplot[mark=triangle*,opacity=0.8,mark options={black,fill=red},mark size=5pt]
	coordinates {
(1,-1.299)
(2,-1.169)
(3,-0.912)
(4,-0.497)
(5,0.112)
(6,0.692)
(7,1.461)
	};\addlegendentry{$r=1$}%
	\addplot[mark=diamond*,opacity=0.8,mark options={black,fill=blue},mark size=5pt]
	coordinates {
(1,-1.297)
(2,-1.113)
(3,-0.748)
(4,-0.237)
(5,0.496)
(6,1.105)
(7,1.934)

	};\addlegendentry{$r=2$}%
	\addplot[mark=*,opacity=0.8,mark options={black,fill=green},mark size=3pt]
	coordinates {
(1,-1.239)
(2,-1.025)
(3,-0.589)
(4,-0.007)
(5,0.782)
(6,1.428)
(7,2.288)
	};\addlegendentry{$r=3$}%
	
	\addplot[mark=square*,opacity=0.8,mark options={black,fill=yellow},mark size=3pt]
	coordinates {
(1,-1.19)
(2,-0.938)
(3,-0.484)
(4,0.194)
(5,1.037)
(6,1.694)
(7,2.578)

	};\addlegendentry{$r=4$}%
	\end{groupplot}
	\node at ($(group c1r2) + (6.0cm,-4.7cm)$) {\ref{grouplegend}}; 
	\end{tikzpicture}
    \vspace{-20pt}
    \caption{Average runtime of \textit{FEATHER} as a function of node and edge count, number of features and characteristic function evaluation points. The mean runtimes were calculated from 10 repetitions using synthetic Erdős Rényi graphs. }\label{fig:feather_runtime}
	\vspace{-5pt}
\end{figure}
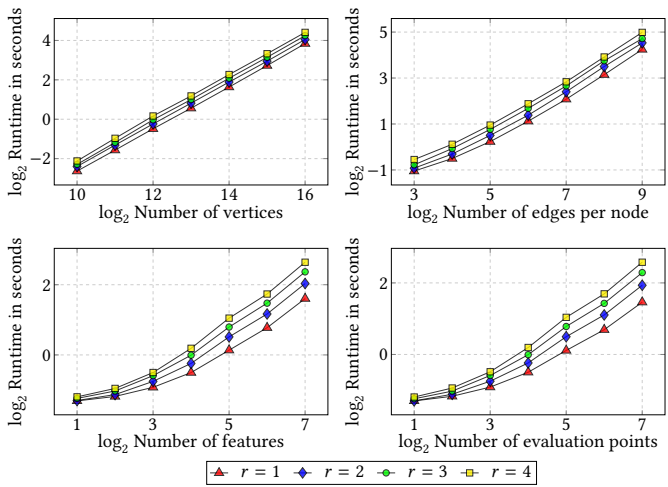

%% file: sections/conclusions.tex
\section{Conclusions and future directions}\label{sec:feather_conclusions}
We presented a general notion of characteristic functions defined on attributed graphs. We discussed a specific instance of these -- the $r$-scale random walk weighted characteristic function. We proposed \textit{FEATHER} an efficient algorithm to calculate this characteristic function efficiently on large attributed graphs in linear time to create Euclidean vector space representations of nodes. We proved that \textit{FEATHER} is robust to data corruption and that isomorphic graphs have the same vector space representations. We have shown that \textit{FEATHER} can be interpreted as the forward pass of a neural network and can be used as a differentiable building block for parametric classifiers.

We demonstrated on various real world node and graph classification datasets that \textit{FEATHER} variants are competitive with comparable embedding and graph neural network models. Our transfer learning results support that \textit{FEATHER} models are able to efficiently and robustly transfer knowledge from one graph to another one. The sensitivity analysis of characteristic function based models and node classification results highlight that supervised \textit{FEATHER} models have an edge compared to unsupervised representation creation with characteristic functions. Furthermore, runtime experiments presented show that our proposed algorithm scales linearly with the input size such as number of nodes, edges and features in practice.

As a future direction we would like to point out that the forward pass of the \textit{FEATHER} algorithm could be incorporated in temporal, multiplex and heterogeneous graph neural network models to serve as a multi-scale vertex feature extraction block. Moreover, one could define a characteristic function based node feature pooling where the node feature aggregation is a learned permutation invariant function. Finally, our evaluation of the proposed algorithms was limited to social networks and web graphs -- testing it on biological networks and other types of datasets could be an important further direction.

%% file: main.bbl

\begin{thebibliography}{48}


\ifx \showCODEN    \undefined \def \showCODEN     #1{\unskip}     \fi
\ifx \showDOI      \undefined \def \showDOI       #1{#1}\fi
\ifx \showISBNx    \undefined \def \showISBNx     #1{\unskip}     \fi
\ifx \showISBNxiii \undefined \def \showISBNxiii  #1{\unskip}     \fi
\ifx \showISSN     \undefined \def \showISSN      #1{\unskip}     \fi
\ifx \showLCCN     \undefined \def \showLCCN      #1{\unskip}     \fi
\ifx \shownote     \undefined \def \shownote      #1{#1}          \fi
\ifx \showarticletitle \undefined \def \showarticletitle #1{#1}   \fi
\ifx \showURL      \undefined \def \showURL       {\relax}        \fi
\providecommand\bibfield[2]{#2}
\providecommand\bibinfo[2]{#2}
\providecommand\natexlab[1]{#1}
\providecommand\showeprint[2][]{arXiv:#2}

\bibitem[\protect\citeauthoryear{Abu-El-Haija, Perozzi, Kapoor, Alipourfard,
  Lerman, Harutyunyan, Steeg, and Galstyan}{Abu-El-Haija et~al\mbox{.}}{2019}]%
        {mixhop}
\bibfield{author}{\bibinfo{person}{Sami Abu-El-Haija}, \bibinfo{person}{Bryan
  Perozzi}, \bibinfo{person}{Amol Kapoor}, \bibinfo{person}{Nazanin
  Alipourfard}, \bibinfo{person}{Kristina Lerman}, \bibinfo{person}{Hrayr
  Harutyunyan}, \bibinfo{person}{Greg~Ver Steeg}, {and} \bibinfo{person}{Aram
  Galstyan}.} \bibinfo{year}{2019}\natexlab{}.
\newblock \showarticletitle{{M}ix{H}op: Higher-Order Graph Convolutional
  Architectures via Sparsified Neighborhood Mixing}. In
  \bibinfo{booktitle}{\emph{International Conference on Machine Learning}}.
\newblock


\bibitem[\protect\citeauthoryear{Ahmed, Rossi, Lee, Willke, Zhou, Kong, and
  Eldardiry}{Ahmed et~al\mbox{.}}{2019}]%
        {role2vec}
\bibfield{author}{\bibinfo{person}{Nesreen~K Ahmed}, \bibinfo{person}{Ryan~A
  Rossi}, \bibinfo{person}{John~Boaz Lee}, \bibinfo{person}{Theodore~L Willke},
  \bibinfo{person}{Rong Zhou}, \bibinfo{person}{Xiangnan Kong}, {and}
  \bibinfo{person}{Hoda Eldardiry}.} \bibinfo{year}{2019}\natexlab{}.
\newblock \showarticletitle{role2vec: Role-based network embeddings}. In
  \bibinfo{booktitle}{\emph{Proc. DLG KDD}}.
\newblock


\bibitem[\protect\citeauthoryear{Cao, Lu, and Xu}{Cao et~al\mbox{.}}{2015}]%
        {grarep}
\bibfield{author}{\bibinfo{person}{Shaosheng Cao}, \bibinfo{person}{Wei Lu},
  {and} \bibinfo{person}{Qiongkai Xu}.} \bibinfo{year}{2015}\natexlab{}.
\newblock \showarticletitle{Grarep: Learning graph representations with global
  structural information}. In \bibinfo{booktitle}{\emph{Proceedings of the 24th
  ACM international on conference on information and knowledge management}}.
  ACM, \bibinfo{pages}{891--900}.
\newblock


\bibitem[\protect\citeauthoryear{Chen and Koga}{Chen and Koga}{2019}]%
        {gl2vec}
\bibfield{author}{\bibinfo{person}{Hong Chen} {and} \bibinfo{person}{Hisashi
  Koga}.} \bibinfo{year}{2019}\natexlab{}.
\newblock \showarticletitle{GL2vec: Graph Embedding Enriched by Line Graphs
  with Edge Features}. In \bibinfo{booktitle}{\emph{International Conference on
  Neural Information Processing}}. Springer, \bibinfo{pages}{3--14}.
\newblock


\bibitem[\protect\citeauthoryear{Chiang, Liu, Si, Li, Bengio, and Hsieh}{Chiang
  et~al\mbox{.}}{2019}]%
        {clustergcn}
\bibfield{author}{\bibinfo{person}{Wei-Lin Chiang}, \bibinfo{person}{Xuanqing
  Liu}, \bibinfo{person}{Si Si}, \bibinfo{person}{Yang Li},
  \bibinfo{person}{Samy Bengio}, {and} \bibinfo{person}{Cho-Jui Hsieh}.}
  \bibinfo{year}{2019}\natexlab{}.
\newblock \showarticletitle{Cluster-GCN: An Efficient Algorithm for Training
  Deep and Large Graph Convolutional Networks}. In
  \bibinfo{booktitle}{\emph{International Conference on Knowledge Discovery and
  Data Mining}}.
\newblock


\bibitem[\protect\citeauthoryear{Cohen, Delling, Pajor, and Werneck}{Cohen
  et~al\mbox{.}}{2014}]%
        {cohen2014distance}
\bibfield{author}{\bibinfo{person}{Edith Cohen}, \bibinfo{person}{Daniel
  Delling}, \bibinfo{person}{Thomas Pajor}, {and} \bibinfo{person}{Renato~F
  Werneck}.} \bibinfo{year}{2014}\natexlab{}.
\newblock \showarticletitle{{Distance-Based Influence in Networks: Computation
  and Maximization}}.
\newblock \bibinfo{journal}{\emph{arXiv preprint arXiv:1410.6976}}
  (\bibinfo{year}{2014}).
\newblock


\bibitem[\protect\citeauthoryear{DasGupta}{DasGupta}{2011}]%
        {lukacs}
\bibfield{author}{\bibinfo{person}{Anirban DasGupta}.}
  \bibinfo{year}{2011}\natexlab{}.
\newblock \bibinfo{booktitle}{\emph{Characteristic Functions and
  Applications}}.
\newblock \bibinfo{pages}{293--322}.
\newblock


\bibitem[\protect\citeauthoryear{de~Lara and Edouard}{de~Lara and
  Edouard}{2018}]%
        {sf}
\bibfield{author}{\bibinfo{person}{Nathan de Lara} {and}
  \bibinfo{person}{Pineau Edouard}.} \bibinfo{year}{2018}\natexlab{}.
\newblock \showarticletitle{A simple baseline algorithm for graph
  classification}. In \bibinfo{booktitle}{\emph{Advances in Neural Information
  Processing Systems}}.
\newblock


\bibitem[\protect\citeauthoryear{Defazio, Bach, and Lacoste-Julien}{Defazio
  et~al\mbox{.}}{2014}]%
        {saga}
\bibfield{author}{\bibinfo{person}{Aaron Defazio}, \bibinfo{person}{Francis
  Bach}, {and} \bibinfo{person}{Simon Lacoste-Julien}.}
  \bibinfo{year}{2014}\natexlab{}.
\newblock \showarticletitle{SAGA: A fast incremental gradient method with
  support for non-strongly convex composite objectives}. In
  \bibinfo{booktitle}{\emph{Advances in neural information processing
  systems}}. \bibinfo{pages}{1646--1654}.
\newblock


\bibitem[\protect\citeauthoryear{Donnat, Zitnik, Hallac, and Leskovec}{Donnat
  et~al\mbox{.}}{2018}]%
        {graphwave}
\bibfield{author}{\bibinfo{person}{Claire Donnat}, \bibinfo{person}{Marinka
  Zitnik}, \bibinfo{person}{David Hallac}, {and} \bibinfo{person}{Jure
  Leskovec}.} \bibinfo{year}{2018}\natexlab{}.
\newblock \showarticletitle{Learning structural node embeddings via diffusion
  wavelets}. In \bibinfo{booktitle}{\emph{Proceedings of the 24th ACM SIGKDD
  International Conference on Knowledge Discovery \& Data Mining}}. ACM,
  \bibinfo{pages}{1320--1329}.
\newblock


\bibitem[\protect\citeauthoryear{Fey and Lenssen}{Fey and Lenssen}{2019}]%
        {pytorch_geometric}
\bibfield{author}{\bibinfo{person}{Matthias Fey} {and} \bibinfo{person}{Jan~E.
  Lenssen}.} \bibinfo{year}{2019}\natexlab{}.
\newblock \showarticletitle{Fast Graph Representation Learning with {PyTorch
  Geometric}}. In \bibinfo{booktitle}{\emph{ICLR Workshop on Representation
  Learning on Graphs and Manifolds}}.
\newblock


\bibitem[\protect\citeauthoryear{Gao, Wolf, and Hirn}{Gao
  et~al\mbox{.}}{2019}]%
        {geoscattering}
\bibfield{author}{\bibinfo{person}{Feng Gao}, \bibinfo{person}{Guy Wolf}, {and}
  \bibinfo{person}{Matthew Hirn}.} \bibinfo{year}{2019}\natexlab{}.
\newblock \showarticletitle{Geometric Scattering for Graph Data Analysis}. In
  \bibinfo{booktitle}{\emph{Proceedings of the 36th International Conference on
  Machine Learning}}, Vol.~\bibinfo{volume}{97}. \bibinfo{pages}{2122--2131}.
\newblock


\bibitem[\protect\citeauthoryear{Gao and Ji}{Gao and Ji}{2019}]%
        {topkpooling}
\bibfield{author}{\bibinfo{person}{Hongyang Gao} {and}
  \bibinfo{person}{Shuiwang Ji}.} \bibinfo{year}{2019}\natexlab{}.
\newblock \showarticletitle{Graph {U-nets}}. In
  \bibinfo{booktitle}{\emph{Proceedings of The 36th International Conference on
  Machine Learning}}.
\newblock


\bibitem[\protect\citeauthoryear{G{\"a}rtner, Flach, and Wrobel}{G{\"a}rtner
  et~al\mbox{.}}{2003}]%
        {rwkernel}
\bibfield{author}{\bibinfo{person}{Thomas G{\"a}rtner}, \bibinfo{person}{Peter
  Flach}, {and} \bibinfo{person}{Stefan Wrobel}.}
  \bibinfo{year}{2003}\natexlab{}.
\newblock \showarticletitle{On graph kernels: Hardness results and efficient
  alternatives}.
\newblock In \bibinfo{booktitle}{\emph{Learning theory and kernel machines}}.
  \bibinfo{publisher}{Springer}, \bibinfo{pages}{129--143}.
\newblock


\bibitem[\protect\citeauthoryear{Grover and Leskovec}{Grover and
  Leskovec}{2016}]%
        {node2vec}
\bibfield{author}{\bibinfo{person}{Aditya Grover} {and} \bibinfo{person}{Jure
  Leskovec}.} \bibinfo{year}{2016}\natexlab{}.
\newblock \showarticletitle{node2vec: Scalable feature learning for networks}.
  In \bibinfo{booktitle}{\emph{Proceedings of the 22nd ACM SIGKDD international
  conference on Knowledge discovery and data mining}}.
  \bibinfo{pages}{855--864}.
\newblock


\bibitem[\protect\citeauthoryear{Halko, Martinsson, and Tropp}{Halko
  et~al\mbox{.}}{2011}]%
        {truncatedsvd}
\bibfield{author}{\bibinfo{person}{N. Halko}, \bibinfo{person}{P.~G.
  Martinsson}, {and} \bibinfo{person}{J.~A. Tropp}.}
  \bibinfo{year}{2011}\natexlab{}.
\newblock \showarticletitle{Finding Structure with Randomness: Probabilistic
  Algorithms for Constructing Approximate Matrix Decompositions}.
\newblock \bibinfo{journal}{\emph{SIAM Rev.}} \bibinfo{volume}{53},
  \bibinfo{number}{2} (\bibinfo{year}{2011}), \bibinfo{pages}{217–288}.
\newblock


\bibitem[\protect\citeauthoryear{Hamilton, Ying, and Leskovec}{Hamilton
  et~al\mbox{.}}{2017}]%
        {graphsage}
\bibfield{author}{\bibinfo{person}{Will Hamilton}, \bibinfo{person}{Zhitao
  Ying}, {and} \bibinfo{person}{Jure Leskovec}.}
  \bibinfo{year}{2017}\natexlab{}.
\newblock \showarticletitle{Inductive Representation Learning on Large Graphs}.
  In \bibinfo{booktitle}{\emph{Advances in Neural Information Processing
  Systems}}.
\newblock


\bibitem[\protect\citeauthoryear{Henderson, Gallagher, Eliassi-Rad, Tong, Basu,
  Akoglu, Koutra, Faloutsos, and Li}{Henderson et~al\mbox{.}}{2012}]%
        {rolx}
\bibfield{author}{\bibinfo{person}{Keith Henderson}, \bibinfo{person}{Brian
  Gallagher}, \bibinfo{person}{Tina Eliassi-Rad}, \bibinfo{person}{Hanghang
  Tong}, \bibinfo{person}{Sugato Basu}, \bibinfo{person}{Leman Akoglu},
  \bibinfo{person}{Danai Koutra}, \bibinfo{person}{Christos Faloutsos}, {and}
  \bibinfo{person}{Lei Li}.} \bibinfo{year}{2012}\natexlab{}.
\newblock \showarticletitle{Rolx: structural role extraction \& mining in large
  graphs}. In \bibinfo{booktitle}{\emph{Proceedings of the 18th ACM SIGKDD
  international conference on Knowledge discovery and data mining}}.
  \bibinfo{pages}{1231--1239}.
\newblock


\bibitem[\protect\citeauthoryear{Horv{\'a}th, G{\"a}rtner, and
  Wrobel}{Horv{\'a}th et~al\mbox{.}}{2004}]%
        {cyclickernel}
\bibfield{author}{\bibinfo{person}{Tam{\'a}s Horv{\'a}th},
  \bibinfo{person}{Thomas G{\"a}rtner}, {and} \bibinfo{person}{Stefan Wrobel}.}
  \bibinfo{year}{2004}\natexlab{}.
\newblock \showarticletitle{Cyclic pattern kernels for predictive graph
  mining}. In \bibinfo{booktitle}{\emph{Proceedings of the tenth ACM SIGKDD
  international conference on Knowledge discovery and data mining}}.
  \bibinfo{pages}{158--167}.
\newblock


\bibitem[\protect\citeauthoryear{Karypis and Kumar}{Karypis and Kumar}{1998}]%
        {metis}
\bibfield{author}{\bibinfo{person}{George Karypis} {and} \bibinfo{person}{Vipin
  Kumar}.} \bibinfo{year}{1998}\natexlab{}.
\newblock \showarticletitle{A fast and high quality multilevel scheme for
  partitioning irregular graphs}.
\newblock \bibinfo{journal}{\emph{SIAM Journal on scientific Computing}}
  \bibinfo{volume}{20}, \bibinfo{number}{1} (\bibinfo{year}{1998}),
  \bibinfo{pages}{359--392}.
\newblock


\bibitem[\protect\citeauthoryear{Kingma and Ba}{Kingma and Ba}{2015}]%
        {adam}
\bibfield{author}{\bibinfo{person}{Diederik Kingma} {and}
  \bibinfo{person}{Jimmy Ba}.} \bibinfo{year}{2015}\natexlab{}.
\newblock \showarticletitle{Adam: A Method for Stochastic Optimization}. In
  \bibinfo{booktitle}{\emph{International Conference on Learning
  Representations}}.
\newblock


\bibitem[\protect\citeauthoryear{Kipf and Welling}{Kipf and Welling}{2017}]%
        {gcn}
\bibfield{author}{\bibinfo{person}{Thomas~N. Kipf} {and} \bibinfo{person}{Max
  Welling}.} \bibinfo{year}{2017}\natexlab{}.
\newblock \showarticletitle{Semi-Supervised Classification with Graph
  Convolutional Networks}. In \bibinfo{booktitle}{\emph{International
  Conference on Learning Representations}}.
\newblock


\bibitem[\protect\citeauthoryear{Klicpera, Bojchevski, and
  G{\"u}nnemann}{Klicpera et~al\mbox{.}}{2019}]%
        {appnp}
\bibfield{author}{\bibinfo{person}{Johannes Klicpera},
  \bibinfo{person}{Aleksandar Bojchevski}, {and} \bibinfo{person}{Stephan
  G{\"u}nnemann}.} \bibinfo{year}{2019}\natexlab{}.
\newblock \showarticletitle{Predict then Propagate: Graph Neural Networks meet
  Personalized PageRank}. In \bibinfo{booktitle}{\emph{International Conference
  on Learning Representations}}.
\newblock


\bibitem[\protect\citeauthoryear{Lee, Lee, and Kang}{Lee et~al\mbox{.}}{2019}]%
        {sagpooling}
\bibfield{author}{\bibinfo{person}{Junhyun Lee}, \bibinfo{person}{Inyeop Lee},
  {and} \bibinfo{person}{Jaewoo Kang}.} \bibinfo{year}{2019}\natexlab{}.
\newblock \showarticletitle{Self-Attention Graph Pooling}. In
  \bibinfo{booktitle}{\emph{Proceedings of the 36th International Conference on
  Machine Learning}}.
\newblock


\bibitem[\protect\citeauthoryear{Liao, He, Zhang, and Chua}{Liao
  et~al\mbox{.}}{2018}]%
        {asne}
\bibfield{author}{\bibinfo{person}{Lizi Liao}, \bibinfo{person}{Xiangnan He},
  \bibinfo{person}{Hanwang Zhang}, {and} \bibinfo{person}{Tat-Seng Chua}.}
  \bibinfo{year}{2018}\natexlab{}.
\newblock \showarticletitle{Attributed Social Network Embedding}.
\newblock \bibinfo{journal}{\emph{IEEE Transactions on Knowledge and Data
  Engineering}} \bibinfo{volume}{30}, \bibinfo{number}{12}
  (\bibinfo{year}{2018}), \bibinfo{pages}{2257--2270}.
\newblock


\bibitem[\protect\citeauthoryear{Narayanan, Chandramohan, Venkatesan, Chen, and
  Liu}{Narayanan et~al\mbox{.}}{2017}]%
        {graph2vec}
\bibfield{author}{\bibinfo{person}{Annamalai Narayanan},
  \bibinfo{person}{Mahinthan Chandramohan}, \bibinfo{person}{Rajasekar
  Venkatesan}, \bibinfo{person}{Lihui Chen}, {and} \bibinfo{person}{Yang Liu}.}
  \bibinfo{year}{2017}\natexlab{}.
\newblock \showarticletitle{graph2vec: Learning distributed representations of
  graphs}.
\newblock  (\bibinfo{year}{2017}).
\newblock


\bibitem[\protect\citeauthoryear{Ou, Cui, Pei, Zhang, and Zhu}{Ou
  et~al\mbox{.}}{2016}]%
        {hope}
\bibfield{author}{\bibinfo{person}{Mingdong Ou}, \bibinfo{person}{Peng Cui},
  \bibinfo{person}{Jian Pei}, \bibinfo{person}{Ziwei Zhang}, {and}
  \bibinfo{person}{Wenwu Zhu}.} \bibinfo{year}{2016}\natexlab{}.
\newblock \showarticletitle{Asymmetric transitivity preserving graph
  embedding}. In \bibinfo{booktitle}{\emph{Proceedings of the 22nd ACM SIGKDD
  international conference on Knowledge discovery and data mining}}.
  \bibinfo{pages}{1105--1114}.
\newblock


\bibitem[\protect\citeauthoryear{Pedregosa, Varoquaux, Gramfort, Michel,
  Thirion, Grisel, Blondel, Prettenhofer, Weiss, Dubourg,
  et~al\mbox{.}}{Pedregosa et~al\mbox{.}}{2011}]%
        {scikit}
\bibfield{author}{\bibinfo{person}{Fabian Pedregosa}, \bibinfo{person}{Ga{\"e}l
  Varoquaux}, \bibinfo{person}{Alexandre Gramfort}, \bibinfo{person}{Vincent
  Michel}, \bibinfo{person}{Bertrand Thirion}, \bibinfo{person}{Olivier
  Grisel}, \bibinfo{person}{Mathieu Blondel}, \bibinfo{person}{Peter
  Prettenhofer}, \bibinfo{person}{Ron Weiss}, \bibinfo{person}{Vincent
  Dubourg}, {et~al\mbox{.}}} \bibinfo{year}{2011}\natexlab{}.
\newblock \showarticletitle{Scikit-learn: Machine learning in Python}.
\newblock \bibinfo{journal}{\emph{Journal of machine learning research}}
  \bibinfo{volume}{12}, \bibinfo{number}{Oct} (\bibinfo{year}{2011}),
  \bibinfo{pages}{2825--2830}.
\newblock


\bibitem[\protect\citeauthoryear{Perozzi, Al-Rfou, and Skiena}{Perozzi
  et~al\mbox{.}}{2014}]%
        {deepwalk}
\bibfield{author}{\bibinfo{person}{Bryan Perozzi}, \bibinfo{person}{Rami
  Al-Rfou}, {and} \bibinfo{person}{Steven Skiena}.}
  \bibinfo{year}{2014}\natexlab{}.
\newblock \showarticletitle{Deepwalk: Online learning of social
  representations}. In \bibinfo{booktitle}{\emph{Proceedings of the 20th ACM
  SIGKDD international conference on Knowledge discovery and data mining}}.
  ACM, \bibinfo{pages}{701--710}.
\newblock


\bibitem[\protect\citeauthoryear{Perozzi, Kulkarni, Chen, and Skiena}{Perozzi
  et~al\mbox{.}}{2017}]%
        {walklets}
\bibfield{author}{\bibinfo{person}{Bryan Perozzi}, \bibinfo{person}{Vivek
  Kulkarni}, \bibinfo{person}{Haochen Chen}, {and} \bibinfo{person}{Steven
  Skiena}.} \bibinfo{year}{2017}\natexlab{}.
\newblock \showarticletitle{Don't Walk, Skip!: online learning of multi-scale
  network embeddings}. In \bibinfo{booktitle}{\emph{Proceedings of the 2017
  IEEE/ACM International Conference on Advances in Social Networks Analysis and
  Mining 2017}}. ACM, \bibinfo{pages}{258--265}.
\newblock


\bibitem[\protect\citeauthoryear{Perozzi and Skiena}{Perozzi and
  Skiena}{2015}]%
        {perozzi2015exact}
\bibfield{author}{\bibinfo{person}{Bryan Perozzi} {and} \bibinfo{person}{Steven
  Skiena}.} \bibinfo{year}{2015}\natexlab{}.
\newblock \showarticletitle{Exact age prediction in social networks}. In
  \bibinfo{booktitle}{\emph{Proceedings of the 24th International Conference on
  World Wide Web}}. \bibinfo{pages}{91--92}.
\newblock


\bibitem[\protect\citeauthoryear{Qiu, Dong, Ma, Li, Wang, and Tang}{Qiu
  et~al\mbox{.}}{2018}]%
        {netmf}
\bibfield{author}{\bibinfo{person}{Jiezhong Qiu}, \bibinfo{person}{Yuxiao
  Dong}, \bibinfo{person}{Hao Ma}, \bibinfo{person}{Jian Li},
  \bibinfo{person}{Kuansan Wang}, {and} \bibinfo{person}{Jie Tang}.}
  \bibinfo{year}{2018}\natexlab{}.
\newblock \showarticletitle{Network embedding as matrix factorization: Unifying
  deepwalk, line, pte, and node2vec}. In \bibinfo{booktitle}{\emph{Proceedings
  of the Eleventh ACM International Conference on Web Search and Data Mining}}.
  ACM, \bibinfo{pages}{459--467}.
\newblock


\bibitem[\protect\citeauthoryear{Rozemberczki, Allen, and Sarkar}{Rozemberczki
  et~al\mbox{.}}{2019a}]%
        {musae}
\bibfield{author}{\bibinfo{person}{Benedek Rozemberczki}, \bibinfo{person}{Carl
  Allen}, {and} \bibinfo{person}{Rik Sarkar}.}
  \bibinfo{year}{2019}\natexlab{a}.
\newblock \showarticletitle{Multi-scale Attributed Node Embedding}.
\newblock \bibinfo{journal}{\emph{arXiv preprint arXiv:1909.13021}}
  (\bibinfo{year}{2019}).
\newblock


\bibitem[\protect\citeauthoryear{Rozemberczki, Davies, Sarkar, and
  Sutton}{Rozemberczki et~al\mbox{.}}{2019b}]%
        {gemsec}
\bibfield{author}{\bibinfo{person}{Benedek Rozemberczki}, \bibinfo{person}{Ryan
  Davies}, \bibinfo{person}{Rik Sarkar}, {and} \bibinfo{person}{Charles
  Sutton}.} \bibinfo{year}{2019}\natexlab{b}.
\newblock \showarticletitle{GEMSEC: Graph Embedding with Self Clustering}. In
  \bibinfo{booktitle}{\emph{Proceedings of the 2019 IEEE/ACM International
  Conference on Advances in Social Networks Analysis and Mining 2019}}. ACM,
  \bibinfo{pages}{65--72}.
\newblock


\bibitem[\protect\citeauthoryear{Rozemberczki, Kiss, and Sarkar}{Rozemberczki
  et~al\mbox{.}}{2020}]%
        {karateclub}
\bibfield{author}{\bibinfo{person}{Benedek Rozemberczki},
  \bibinfo{person}{Oliver Kiss}, {and} \bibinfo{person}{Rik Sarkar}.}
  \bibinfo{year}{2020}\natexlab{}.
\newblock \showarticletitle{{Karate Club: An API Oriented Open-source Python
  Framework for Unsupervised Learning on Graphs}}. In
  \bibinfo{booktitle}{\emph{Proceedings of the 29th ACM International on
  Conference on Information and Knowledge Management (CIKM '20)}}. ACM.
\newblock


\bibitem[\protect\citeauthoryear{Rozemberczki and Sarkar}{Rozemberczki and
  Sarkar}{2018}]%
        {diff2vec}
\bibfield{author}{\bibinfo{person}{Benedek Rozemberczki} {and}
  \bibinfo{person}{Rik Sarkar}.} \bibinfo{year}{2018}\natexlab{}.
\newblock \showarticletitle{Fast Sequence-Based Embedding with Diffusion
  Graphs}. In \bibinfo{booktitle}{\emph{International Workshop on Complex
  Networks}}. Springer, \bibinfo{pages}{99--107}.
\newblock


\bibitem[\protect\citeauthoryear{Shervashidze, Schweitzer, Van~Leeuwen,
  Mehlhorn, and Borgwardt}{Shervashidze et~al\mbox{.}}{2011}]%
        {wlkernel}
\bibfield{author}{\bibinfo{person}{Nino Shervashidze}, \bibinfo{person}{Pascal
  Schweitzer}, \bibinfo{person}{Erik~Jan Van~Leeuwen}, \bibinfo{person}{Kurt
  Mehlhorn}, {and} \bibinfo{person}{Karsten~M Borgwardt}.}
  \bibinfo{year}{2011}\natexlab{}.
\newblock \showarticletitle{Weisfeiler-lehman graph kernels}.
\newblock \bibinfo{journal}{\emph{Journal of Machine Learning Research}}
  \bibinfo{volume}{12}, \bibinfo{number}{77} (\bibinfo{year}{2011}),
  \bibinfo{pages}{2539--2561}.
\newblock


\bibitem[\protect\citeauthoryear{Srivastava, Hinton, Krizhevsky, Sutskever, and
  Salakhutdinov}{Srivastava et~al\mbox{.}}{2014}]%
        {dropout}
\bibfield{author}{\bibinfo{person}{Nitish Srivastava},
  \bibinfo{person}{Geoffrey Hinton}, \bibinfo{person}{Alex Krizhevsky},
  \bibinfo{person}{Ilya Sutskever}, {and} \bibinfo{person}{Ruslan
  Salakhutdinov}.} \bibinfo{year}{2014}\natexlab{}.
\newblock \showarticletitle{Dropout: A Simple Way to Prevent Neural Networks
  from Overfitting}.
\newblock \bibinfo{journal}{\emph{J. Mach. Learn. Res.}} \bibinfo{volume}{15},
  \bibinfo{number}{1} (\bibinfo{year}{2014}), \bibinfo{pages}{1929–1958}.
\newblock


\bibitem[\protect\citeauthoryear{Tang, Qu, Wang, Zhang, Yan, and Mei}{Tang
  et~al\mbox{.}}{2015}]%
        {line}
\bibfield{author}{\bibinfo{person}{Jian Tang}, \bibinfo{person}{Meng Qu},
  \bibinfo{person}{Mingzhe Wang}, \bibinfo{person}{Ming Zhang},
  \bibinfo{person}{Jun Yan}, {and} \bibinfo{person}{Qiaozhu Mei}.}
  \bibinfo{year}{2015}\natexlab{}.
\newblock \showarticletitle{Line: Large-scale information network embedding}.
  In \bibinfo{booktitle}{\emph{Proceedings of the 24th international conference
  on world wide web}}. \bibinfo{pages}{1067--1077}.
\newblock


\bibitem[\protect\citeauthoryear{Tsitsulin, Mottin, Karras, Bronstein, and
  M{\"u}ller}{Tsitsulin et~al\mbox{.}}{2018}]%
        {netlsd}
\bibfield{author}{\bibinfo{person}{Anton Tsitsulin}, \bibinfo{person}{Davide
  Mottin}, \bibinfo{person}{Panagiotis Karras}, \bibinfo{person}{Alexander
  Bronstein}, {and} \bibinfo{person}{Emmanuel M{\"u}ller}.}
  \bibinfo{year}{2018}\natexlab{}.
\newblock \showarticletitle{Netlsd: hearing the shape of a graph}. In
  \bibinfo{booktitle}{\emph{Proceedings of the 24th ACM SIGKDD International
  Conference on Knowledge Discovery \& Data Mining}}.
  \bibinfo{pages}{2347--2356}.
\newblock


\bibitem[\protect\citeauthoryear{Veli{\v{c}}kovi{\'{c}}, Cucurull, Casanova,
  Romero, Li{\`{o}}, and Bengio}{Veli{\v{c}}kovi{\'{c}} et~al\mbox{.}}{2018}]%
        {gat}
\bibfield{author}{\bibinfo{person}{Petar Veli{\v{c}}kovi{\'{c}}},
  \bibinfo{person}{Guillem Cucurull}, \bibinfo{person}{Arantxa Casanova},
  \bibinfo{person}{Adriana Romero}, \bibinfo{person}{Pietro Li{\`{o}}}, {and}
  \bibinfo{person}{Yoshua Bengio}.} \bibinfo{year}{2018}\natexlab{}.
\newblock \showarticletitle{{Graph Attention Networks}}. In
  \bibinfo{booktitle}{\emph{International Conference on Learning
  Representations}}.
\newblock


\bibitem[\protect\citeauthoryear{Verma and Zhang}{Verma and Zhang}{2017}]%
        {fgsd}
\bibfield{author}{\bibinfo{person}{Saurabh Verma} {and} \bibinfo{person}{Zhi-Li
  Zhang}.} \bibinfo{year}{2017}\natexlab{}.
\newblock \showarticletitle{Hunt for the unique, stable, sparse and fast
  feature learning on graphs}. In \bibinfo{booktitle}{\emph{Advances in Neural
  Information Processing Systems}}. \bibinfo{pages}{88--98}.
\newblock


\bibitem[\protect\citeauthoryear{Wu, Souza, Zhang, Fifty, Yu, and
  Weinberger}{Wu et~al\mbox{.}}{2019}]%
        {sgconv}
\bibfield{author}{\bibinfo{person}{Felix Wu}, \bibinfo{person}{Amauri Souza},
  \bibinfo{person}{Tianyi Zhang}, \bibinfo{person}{Christopher Fifty},
  \bibinfo{person}{Tao Yu}, {and} \bibinfo{person}{Kilian Weinberger}.}
  \bibinfo{year}{2019}\natexlab{}.
\newblock \showarticletitle{{Simplifying Graph Convolutional Networks}}. In
  \bibinfo{booktitle}{\emph{International Conference on Machine Learning}}.
\newblock


\bibitem[\protect\citeauthoryear{Yang, Liu, Zhao, Sun, and Chang}{Yang
  et~al\mbox{.}}{2015}]%
        {tadw}
\bibfield{author}{\bibinfo{person}{Cheng Yang}, \bibinfo{person}{Zhiyuan Liu},
  \bibinfo{person}{Deli Zhao}, \bibinfo{person}{Maosong Sun}, {and}
  \bibinfo{person}{Edward Chang}.} \bibinfo{year}{2015}\natexlab{}.
\newblock \showarticletitle{Network representation learning with rich text
  information}. In \bibinfo{booktitle}{\emph{Twenty-Fourth International Joint
  Conference on Artificial Intelligence}}.
\newblock


\bibitem[\protect\citeauthoryear{Yang, Pan, Zhang, Chen, Lian, and Zhang}{Yang
  et~al\mbox{.}}{2018}]%
        {bane}
\bibfield{author}{\bibinfo{person}{Hong Yang}, \bibinfo{person}{Shirui Pan},
  \bibinfo{person}{Peng Zhang}, \bibinfo{person}{Ling Chen},
  \bibinfo{person}{Defu Lian}, {and} \bibinfo{person}{Chengqi Zhang}.}
  \bibinfo{year}{2018}\natexlab{}.
\newblock \showarticletitle{Binarized attributed network embedding}. In
  \bibinfo{booktitle}{\emph{2018 IEEE International Conference on Data Mining
  (ICDM)}}. IEEE, \bibinfo{pages}{1476--1481}.
\newblock


\bibitem[\protect\citeauthoryear{Yang and Yang}{Yang and Yang}{2018}]%
        {tene}
\bibfield{author}{\bibinfo{person}{Shuang Yang} {and} \bibinfo{person}{Bo
  Yang}.} \bibinfo{year}{2018}\natexlab{}.
\newblock \showarticletitle{Enhanced Network Embedding with Text Information}.
  In \bibinfo{booktitle}{\emph{2018 24th International Conference on Pattern
  Recognition (ICPR)}}. IEEE, \bibinfo{pages}{326--331}.
\newblock


\bibitem[\protect\citeauthoryear{Zhang, Yin, Zhu, and Zhang}{Zhang
  et~al\mbox{.}}{2018b}]%
        {sine}
\bibfield{author}{\bibinfo{person}{Daokun Zhang}, \bibinfo{person}{Jie Yin},
  \bibinfo{person}{Xingquan Zhu}, {and} \bibinfo{person}{Chengqi Zhang}.}
  \bibinfo{year}{2018}\natexlab{b}.
\newblock \showarticletitle{SINE: Scalable Incomplete Network Embedding}. In
  \bibinfo{booktitle}{\emph{2018 IEEE International Conference on Data Mining
  (ICDM)}}. IEEE, \bibinfo{pages}{737--746}.
\newblock


\bibitem[\protect\citeauthoryear{Zhang, Cui, Neumann, and Chen}{Zhang
  et~al\mbox{.}}{2018a}]%
        {sortpooling}
\bibfield{author}{\bibinfo{person}{Muhan Zhang}, \bibinfo{person}{Zhicheng
  Cui}, \bibinfo{person}{Marion Neumann}, {and} \bibinfo{person}{Yixin Chen}.}
  \bibinfo{year}{2018}\natexlab{a}.
\newblock \showarticletitle{An End-to-End Deep Learning Architecture for Graph
  Classification}. In \bibinfo{booktitle}{\emph{AAAI}}.
  \bibinfo{pages}{4438--4445}.
\newblock


\end{thebibliography}
